\documentclass[preprint,review,10pt, authoryear]{elsarticle}

\usepackage{graphicx}
\usepackage{amsmath}
\usepackage{amssymb}
\usepackage{amsthm}
\usepackage{booktabs}
\usepackage{setspace}
\usepackage{lineno}

\usepackage[svgnames]{xcolor}
\usepackage{listings}
\usepackage[T1]{fontenc}
\usepackage{cfr-lm}
\usepackage{bm}

\usepackage{lineno}
\usepackage{nicefrac}
\usepackage{microtype}
\usepackage{subcaption}

\usepackage{hyperref}
\hypersetup{%
  linkcolor  = SteelBlue,
  citecolor  = SteelBlue,
  urlcolor   = SteelBlue,
  colorlinks = true,
}

\usepackage{pgfplots}
\DeclareUnicodeCharacter{2212}{−}
\usepgfplotslibrary{groupplots,dateplot}
\usetikzlibrary{patterns,shapes.arrows}
\pgfplotsset{compat=newest}

\usepackage{natbib}

\usepackage{mathtools}  

\journal{arXiv Preprint}

\makeatletter
\def\@author#1{%
    \g@addto@macro\elsprelimauthors{%
     \prelimauthorsep#1%
     \def\prelimauthorsep{\unskip,\space}}%
    \g@addto@macro\elsauthors{\normalsize%
    \def\baselinestretch{1}%
    \upshape\authorsep#1\unskip\textsuperscript{%
      \ifx\@fnmark\@empty\else\unskip\sep\@fnmark\let\sep=,\fi
      \ifx\@corref\@empty\else\unskip\sep\@corref\let\sep=,\fi
      }%
    \def\authorsep{\unskip\space\textcircled{r}\space}%
    \global\let\@fnmark\@empty
    \global\let\@corref\@empty    \global\let\sep\@empty}%
    \@eadauthor={#1}%
    \g@addto@macro\useauthors{#1; }%
}
\makeatother
\begin{document}
\begin{frontmatter}

\title{A Comprehensive Modeling Approach for Crop Yield Forecasts using
AI-based Methods and Crop Simulation Models}



\author{Renato Luiz de Freitas Cunha\fnref{fn1}}
\author{Bruno Silva\corref{cor1}\fnref{fn1}}
\author{Priscilla Barreira Avegliano\fnref{fn1}}
\cortext[cor1]{
  Corresponding author.
}
\fntext[fn1]{
  Author ordering determined randomly.
}
\address{
    \texttt{\{renatoc,sbruno,pba\}@br.ibm.com}\\
    IBM Research
}

\begin{abstract}

  Numerous solutions for yield estimation are either based on data-driven
  models, or on crop-simulation models (CSMs). Researchers tend to build
  data-driven models using nationwide crop information databases provided by
  agencies such as the USDA. On the opposite side of the spectrum, CSMs require
  fine data that may be hard to generalize from a handful of fields. In this
  paper, we propose a comprehensive approach for yield forecasting that combines
  data-driven solutions, crop simulation models, and model surrogates to support
  multiple user-profiles and needs when dealing with crop management
  decision-making. To achieve this goal, we have developed a solution to
  calibrate CSMs at scale, a surrogate model of a CSM assuring faster execution,
  and a neural network-based approach that performs efficient risk assessment in
  such settings.  Our data-driven modeling approach outperforms previous works
  with yield correlation predictions close to 91\%. The crop simulation modeling
  architecture achieved 6\% error; the proposed crop simulation model surrogate
  performs predictions almost 100 times faster than the adopted crop simulator
  with similar accuracy levels.

\end{abstract}

\begin{keyword}
  Yield Forecasting \sep{} Artificial Intelligente 
  \sep{} Crop Simulation Model


\end{keyword}

\end{frontmatter}


\section{Introduction}

Compared to other objectives such as social development and economic growth,
food production received less attention from policymakers~\citep{godfray2014food}
in the last decades.  Due to the volatility of
food prices in recent years, this situation started to change, revealing the
possibility of millions of people facing famine~\citep{bailey2011growing}, with
governments challenged by the task of feeding 9 billion people by 2050.
To reduce the stress on the environment, we face an imperative to improve crop
productivity with current farmlands, as opposed to simply increasing planted
areas to fulfill the upcoming rise in food demand~\citep{keating2014food}.
To achieve this goal, food security planners should have a comprehensive
understanding of crop yield under different weather, soil, management, and
plant genetics conditions. In this sense, crop yield models are reliable tools
for crop yield estimates~\citep{lobell2009crop}.


Data-driven yield forecast solutions rely on machine learning methods, yields
from previous seasons, weather, soil and other related parameters to estimate the
crop yield~\citep{vanklompenburg2020105709}. The main advantage of such methods
is the possibility of evaluating a given crop metric (e.g., leaf area index)
with a reduced number of input parameters, according to the availability of
data from previous seasons. By using these methods, one is able to forecast the
yield for a large region with reduced consumption of computational resources.
Data-driven solutions have poor performance when predicting yield for scenarios
that differ too much from previously-seen ones~\citep{solomatine2009data}. On
the other hand, crop simulation models perform well in situations when no
previous data is available, as they consider the biological phenomena and
interactions with the environment governing the development of
crops~\citep{akinseye2017assessing}.  Moreover, in both types of models, to have
a clear view of risks associated with crop development, one needs to perform
multiple executions of either model, which is costly and time-consuming.

Crop simulation models work with a detailed representation of plant physiology
and the biological response to weather, soil, genetic and management inputs.
These models output multiple metrics, such as yield, above-ground biomass, leaf
area index, physiological maturity, and seed density. Another important feature
of crop simulation models is their ability to compute output metrics along
time~\citep{hoogenboom2019dssat}, as opposed to only at the end of the
simulation.  The main drawback of this approach is the large amount of data
required to execute a single simulation, the time to complete it, and the
computation resources required to evaluate multiple farming management
scenarios. Therefore, large-scale crop simulation models (e.g., at a global
scale) require huge amounts of computing power. State of the art solutions for
large regions are limited to data-driven models, as crop simulation models
depend on more data to perform predictions.  There is a gap in crop simulation
modeling infrastructure for large regions.  This work aims at filling this gap
by presenting a system for large-scale crop calibration and simulation
environment.

Crop yield prediction is a key challenge in the field of precision agriculture
with many solutions and models proposed so far~\citep{vanklompenburg2020105709}.
This is not a trivial task as it requires multiple data sources such as climate,
weather, soil, and farming management. \citet{nguyen2019spatial} presented a
multi-task learning approach to estimate cotton yield production using a
within-field method. The authors used an approach that leverages production from
previous years to estimate productivity and claim the work is the first attempt
to predict fine-grain cotton yield using a Multi-task learning mechanism. Other
works assume homogeneity within crop fields, the authors of this paper consider
model spatial variations in a given area for soil, climate, tillage, irrigation
conditions and the neighbors' potential correlation in yield assessment.  

\citet{kuwata2015estimating} developed a model using SVR (Support Vector
Regression) and deep learning to predict yield. In the study, the authors
employed NDVI (Normalized Vegetation Index), APAR (Absorbed Photosynthetically
Active Radiation), canopy surface temperature, and water stress as input
features.  The work presented by \citet{cunha2020estimating} employed deep
learning to predict the yield for five major Brazilian crops (Corn, Cotton, Rice,
Soybean, Sugarcane). The authors show that agriculture stakeholders can get
insights into potential productivity even before planting to improve the farming
decision process. \citet{kim2016machine} created a machine learning model to
estimate corn yield in Iowa using different methods such as SVM (Support Vector
Machines), RF (Random Forest), ERT (Extremely Randomized Trees) and DL (Deep
Learning). The authors find out that DL provides better results when compared to
the other machine learning methods. 

\citet{zhang2021408} provided a blended process-based and remote sensing-driven
crop yield model for maize in Northeast China. They leverage remote sensing
strategies to improve crop simulation model estimates that suffer from sparse
weather measurement stations and limited field management information. To
integrate the two modeling approaches, the authors created a brand-new crop
yield estimation model called PRYM-Maize.  \citet{vanklompenburg2020105709}
presents a systematic literature review of machine learning approaches used for
crop yield estimation. According to their analysis, the most used features in
the observed literature are temperature, rainfall, and soil type, and the most
applied algorithm is Artificial Neural Networks in these models. The previous
works concentrate their efforts on machine learning-based crop yield estimation
using different AI methods including Linear Regression, Random Forests, Decision
trees, and Deep Artificial Neural Networks. Although these methods work very
well in predicting yield for scenarios similar to past ones, they fail to
estimate production for conditions distant from those present in the yield and
features database.  Our approach enables users to predict yield for
unseen scenarios by leveraging crop simulation models which can emulate the
physiological response to external events (e.g., weather, soil, and management).

\citet{battisti2017inter} evaluated five different
crop models and their ensemble for soybean yield estimation in
Southern Brazil. The evaluated models include FAO – Agroecological Zone,
AQUACROP, DSSAT CSM CROPGRO Soybean, APSIM Soybean, and MONICA\@. They
evaluated several crop variables including grain yield, crop phases, harvest
index, total above-ground biomass, and leaf area index. The results showed
that ensembles of calibrated models were more accurate than any single crop
simulation model. \citet{rodriguez2018predicting} combined crop
simulation models and seasonal climate forecasting systems to prescribe
combinations of genetics (G), and agronomic management (crop designs
(GxM)) for sorghum fields. Their results showed that optimum crops could be
achieved using the aforementioned models. 
\citet{akinseye2017assessing} studied the performance
of three crop simulation models (APSIM, DSSAT, and Samara) for predicting
sorghum development in West Africa. Their results confirm the capacity of
each model to predict the growth and development of different varieties with
multiple photoperiod sensitivities. The authors also show that Samara
outperforms LAI dynamics and early biomass production estimates during
the vegetative phase when compared to APSIM and DSSAT.
Global warming is an important issue that should be addressed when political
leaders create food security plans for their population. In that regard,
\citet{sultan2019evidence} employed a large ensemble of historical climate
simulation and process-based crop models (SARRA-H and CYGMA) to estimate the
effects of climate change on crop production in West Africa. They have found
a temperature increase of $1^{\circ} \mathrm{C}$ in that region and that led
to 10-20\% millet and 5-15\% sorghum yield reduction. 

The previous works using crop simulation models for enabling users to evaluate
yield or other crop outputs are focused on specific conditions and relatively
small regions. Our work employs High-Performance Computing (HPC) methods to
enable large area model simulation and calibration.


In this work, we propose an approach that provides yield prediction for large
areas using a combination of data-driven yield models and crop simulation
models. Our solution takes the benefits from crop simulation models to promote
farming management decision-making and the possibility of scalable yield
prediction for large areas supported by data-driven yield models. We also
created machine learning-based surrogates to mimic the behavior of crop
simulation models and provide quick results for the evaluation of large regions
with different farming management approaches. In summary, our main
contributions are:

\begin{itemize}

  \item A comprehensive data-driven modeling approach to evaluate crop yield
  using weather, soil and seasonal weather forecasts data. The proposed method
  can predict the potential crop productivity before the actual planting
  process. We also produce crop yield parametric probability distributions
  for risk assessment purposes (Section~\ref{sec:neural-network}).
  
  \item An infrastructure for large-scale crop simulation calibration and
  execution. It supports wide-area evaluations and what-if analysis for
  farming decision-making and optimization support
  (Section~\ref{sec:crop-simulation-model}).

  \item An automatic tool for creating crop simulation model surrogates
to reduce the computational costs of executing the
  actual simulation models and optimizing farming management decisions
  (Section~\ref{sec:surrogate}).

\end{itemize}

Table~\ref{ref:comparisson-table} presents comparatively a summary of the
requirements and performance of the different methods aforementioned. The most
suitable model to be used in a decision-making process should make a tradeoff
between data availability for model creation and execution and available
computational resources.  According to the table, surrogate crop simulation
seems to be the most useful method as it can evaluate field-level data at low
computational requirements.  However, it requires the execution of a crop
simulation model and training of the surrogate using the underlying results. 

\begin{table}[]
\centering
\label{ref:comparisson-table}
	\caption{Comparisson table of presented models}
\resizebox{\textwidth}{!} {%
\begin{tabular}{@{}lcccc@{}}
\toprule
\textbf{Crop model}                   & \textbf{\begin{tabular}[c]{@{}c@{}}Dataset size \\ for model creation\end{tabular}} & \textbf{\begin{tabular}[c]{@{}c@{}}Dataset size \\ for model execution\end{tabular}} & \textbf{\begin{tabular}[c]{@{}c@{}}Computing resources\\  for execution\end{tabular}} & \multicolumn{1}{l}{\textbf{Accuracy}} \\ \midrule
\textit{Data-driven yield prediction} & high                                                                                 & coarse-grained data                                                                  & low                                                                                   & medium                                \\
\textit{Large-scale crop simulation}  & proportional to the evaluation area                                                  & field level fine-grained daily data                                                  & high                                                                                  & high                                  \\
\textit{Surrogate crop simulation}    & proportional to the evaluation area                                                  & field level fine-grained daily data                                                  & low                                                                                   & high                                  \\ \bottomrule
\end{tabular}
}
\end{table}

\section{Method Overview}

\begin{figure}[!h]
    \centering
    \includegraphics[width=0.8\linewidth]{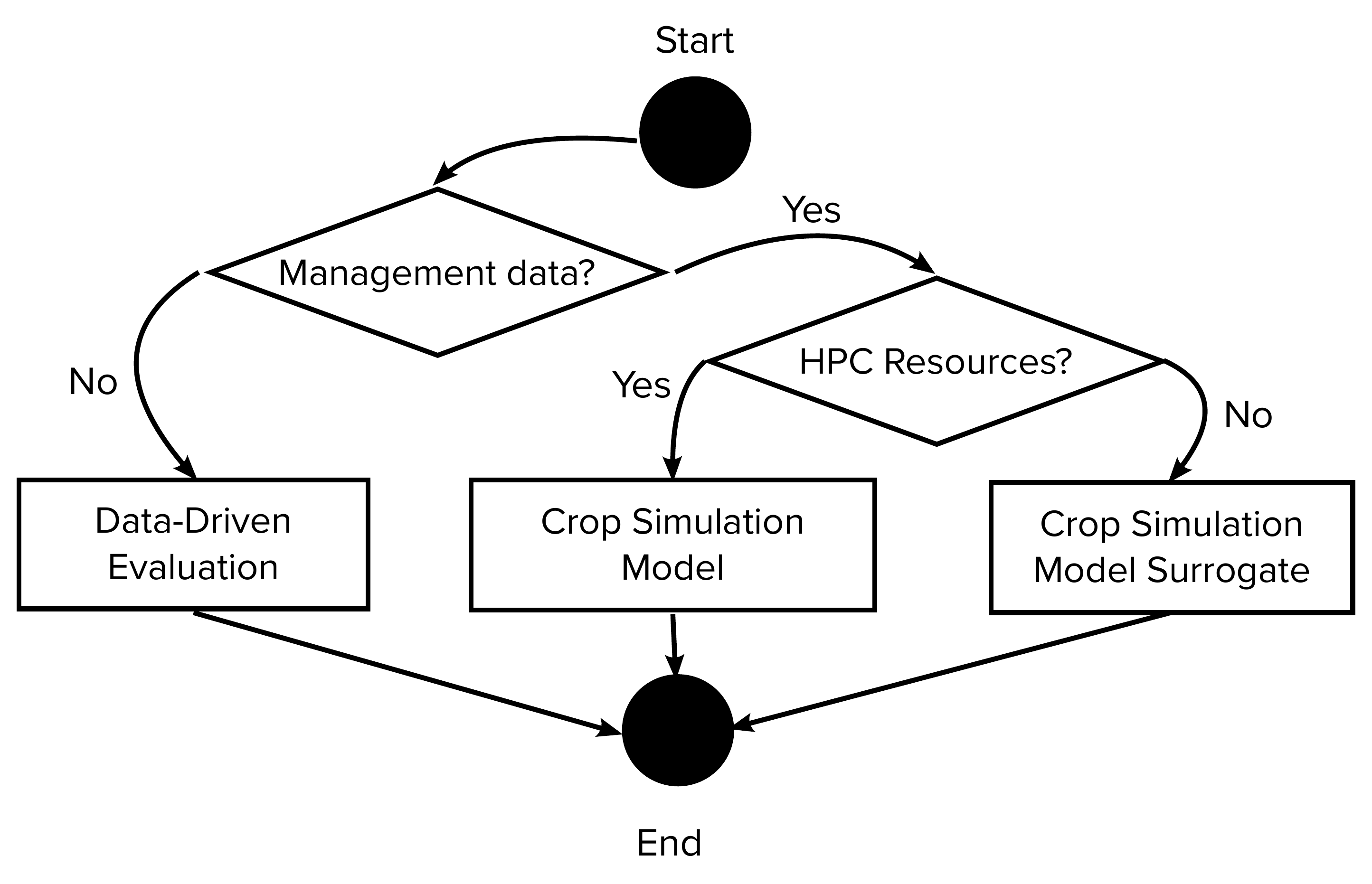}
    \caption{
      The proposed crop yield estimation approach involves three different
      types of user needs. We present a data-driven yield estimation method
      that does not require fine-grained farming management/genetics data.
      There is also a crop simulation model-based approach in which users can
      have farming management insights using a supporting computing
      infrastructure. The last crop yield estimation method employs a crop
      simulation model surrogate to support farming management decisions
      using less computing resources.
    }\label{fig:methodology}
\end{figure}

In this paper, we propose a method to provide crop yield estimates for
different user needs. The method capabilities vary according to the
availability of farming management data and computing resources for system
calibration and evaluation. For instance, food security planners may want
to evaluate the potential yield for a large area without knowing
details about each farmer's planting and cropping strategies.
On the other hand,
a different kind of farming decision-maker may want to assess the potential
yield for a local crop using different types of seeds, planting period, and
management decisions (such as irrigation or fertilization actions).  The
possible system usages are detailed as follows and presented in
Figure~\ref{fig:methodology}.

\textbf{Data-Driven Evaluation}. This evaluation is suited for users who want
to evaluate large areas with different geographical characteristics and do
not have detailed information about where, when, and how a given crop was
planted (i.e., farming management data). This can be the case of food
security politicians interested in investigating the potential crop yield for
some region that can be a state or even an entire country. In this case, most
of the data is available annually at the county level. Therefore, the model
should predict yield for a given county/year without considering multiple
farming management actions within the evaluated region. For this type of
user, we suggest the utilization of data-driven crop regression models using
seasonal weather forecasts in conjunction with soil, weather, and previous
yield data to predict potential yield for large areas. We explain details
related to our data-driven modelling approach in
Section~\ref{sec:neural-network}.

\textbf{Crop Simulation Model}. For farmers who want to make decisions about
the farming management process, we provide a crop simulation model wrapper
that enables users to perform what-if analysis considering different crop
types (see Section~\ref{sec:crop-simulation-model}). Besides the
user-provided farming management data, the system automatically collects
weather/soil data from third-party services and contains an automatic
calibration engine for finding the genetics crop properties. The calibration
and data-gathering service can be time-consuming depending on the evaluated
area size and the number of farming management variable combinations.
Therefore, sometimes a High-Performance Computing (HPC) infrastructure is
required to use this type of service~\citep{silva2018jobpruner}. To overcome
this issue, we propose a surrogate modeling approach that mimics the behavior
of crop simulation models. It enables users that do not have access to HPC
infrastructures to predict the yield of their crops.

\textbf{Crop Simulation Model Surrogate}. To overcome the possible scenarios with limited
access to HPC structure, we created a surrogate model to mimic the behaviour of a widely adopted 
crop simulator (DSSAT). Our approach relies on a space-filling sampling strategy (quasi-random sampling)
 for collecting examples of the behaviour of our target model (in this work DSSAT, but could be any other
 crop simulator) given a set of values of its input variables, to construct a dataset.
 A neural-network was later trained using the generated dataset to replace
 the target model on tasks that require a high number of executions, such as risk assessment. 
This type of data-driven approach can
save computational resources allowing the decision-making process to execute
in limited resources machinery (e.g., commodity hardware). The details about
the crop simulation model surrogate are presented in
Section~\ref{sec:surrogate}.


\section{Data-Driven Evaluation}\label{sec:neural-network}

In this paper, we extend the model presented by \citet{oliveira2018scalable} and
\citet{cunha2020estimating} by adjusting the formulation of the model such that,
instead of providing point estimates, it provides \emph{distributions} for yield
prediction. Users can leverage this feature to perform risk assessment using
probability distribution functions in the decision-making process for farming
management. The current model employs the following features: daily temperature,
precipitation, solar irradiation, growing degree days (GDD), and soil-related
data.

We used the CHIRPS dataset as a source for monthly precipitation
data~\citep{funk2015climate}. CHIRPS has 0.05$^{\circ}$ resolution and is built
by merging satellite and weather station information. CHIRPS uses satellite data
in three ways: satellite means are used to produce high-resolution rainfall
climatologies, infrared Cold Cloud Duration (CCD) fields are used to estimate
monthly and pentadal rainfall deviation from climatologies. Lastly, satellite
precipitation fields are used to guide interpolation through local distance
decay functions.

\begin{figure*}[ht]
  \centering
  \includegraphics[scale=0.7]{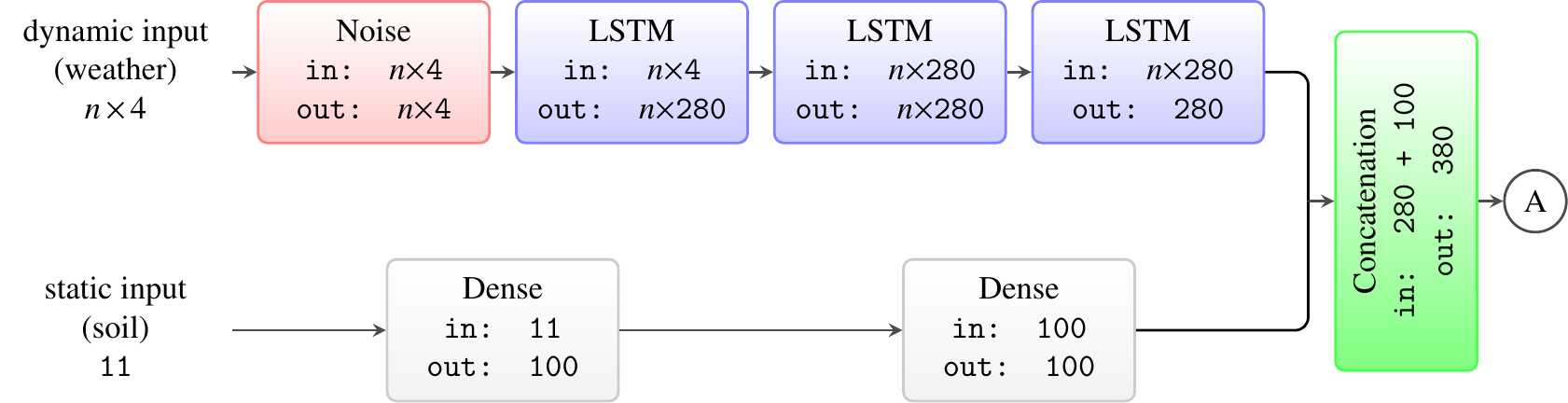}

  \vspace{0.5cm} 

  \includegraphics[scale=0.7]{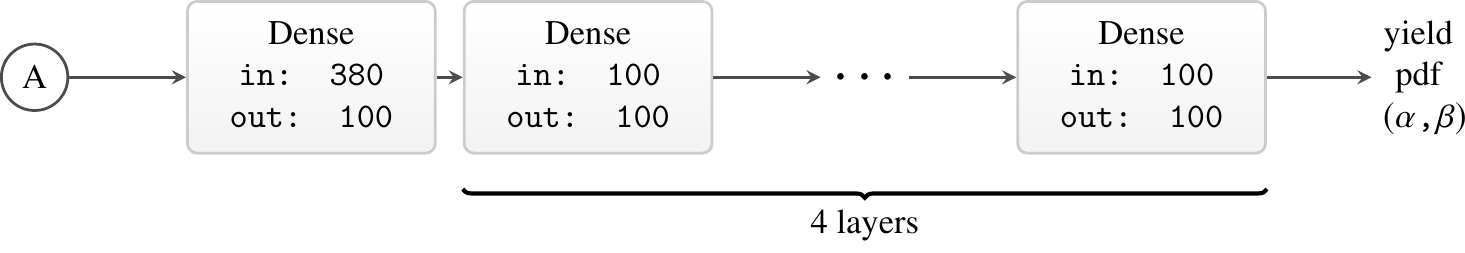}
  \caption{%
    Architecture of the updated model with support for variable time lengths
    in the dynamic (weather) data path. The Red node corresponds
    to a noise generation layer. Blue nodes represent Long Short-Rerm
    Memory (LSTM) recurrent layers. Gray nodes represent dense,
    fully-connected layers. The green node represents a concatenation
    layer, which concatenates the intermediate representations from the
    dynamic and static paths. Numbers below node names represent shapes
    of input and output tensors. For example, $x \times y$ represents a
    $x$ by $y$ matrix, while single numbers represent line vectors. $n$
    corresponds to the crop cycle length (in months). This figure was
    extended from \citet{cunha2020estimating} by adding probability distribution
    parameters $\mu$ and $\sigma$.
  }\label{fig:goronet2}
\end{figure*}

Monthly air temperature data, specifically minimum and maximum temperatures for
each month, were provided by reanalysis datasets from \textsc{era5},
a product developed by the European Centre for Medium-Range Weather
Forecasts (ECMWF)~\citep{dee2011era}. This dataset covers the globe at
a resolution of approximately 80km per pixel and was generated using data
assimilation from multiple sources. We also incorporated GDD as a feature to
account for the influence of temperature on the crops studied. This decision is
justified because different crops have different GDD values. For example, corn
requires around 1600--1770$^\circ$\textsc{c} GDD for achieving full-season
maturity~\citep{neild1987nch}, while soybeans require around
1300--1500$^\circ$\textsc{c} GDD from planting to physiological
maturity~\citep{george1990effect}. In providing this feature, we expect the
model to have more opportunities to learn meaningful mappings from features to
predicted yield.

Soil properties data comes from SoilGrids.org~\citep{hengl2017soilgrids250m},
an open, global soil dataset with a resolution of 250m per pixel which
provides information for clay, silt and sand contents plus fine earth and
coarse fragments bulk density. All this data is available in seven depths ($0$,
$5$, $15$, $30$, $60$, $100$ and $200$cm). SoilGrids data are results of predictions
based on $150000$ soil profiles used for training and $158$ remote sensing-based
soil covariates. These were used to fit an ensemble of random forest,
gradient boosting and multinomial logistic regression models. The model uses
actual yield data which can be private or provided by government agencies
such as IBGE (Brazil Statistics and Geography Bureau)~\citep{ibgepma} and USDA
(United States Department of Agriculture)~\citep{nass2018united}. For each case,
we constructed a dataset that consisted of yields from 2011 to 2018 (inclusive).
The model was trained with data from 2011 to 2017 and evaluated on 2018 data.

Regarding the model's structure (Figure~\ref{fig:goronet2}), it has two
separate data paths (dynamic and static) that merge inside the model. The
rationale behind this design decision is that, in doing so, so-called dynamic
data, such as weather data, can be processed and specialized by different
nodes than the ones focused on static data, such as location and soil data. More specifically,
time-series data such as weather forecasts can be processed by an
LSTM~\citep{hochreiter1997long} nodes, which tend to work well with
time-series data, while soil data can be processed by fully-connected nodes.

Our model expands on the original
model~\citep{oliveira2018scalable,cunha2020estimating} by (i)
adding accumulated GDD as a dynamic feature alongside weather forecasts; (ii)
making the length of the dynamic features dependent on the type of crop being
used by using crop calendars as input; (iii) including an additive zero-centered
Gaussian noise node at the input of the dynamic data path; and (iv) by
formulating the task to predict the parameters of a Normal distribution, as
opposed to predicting only a point estimate. We also improved the original
model~\citep{oliveira2018scalable} by supporting variable window lengths and thus
respecting the planting calendars of different crops. We did this by
incorporating domain knowledge of crop development into our model. This data
comes from crop calendars, which indicate typical planting and harvesting dates
for crops in a given region. In Brazil, this information is published by
\emph{Companhia Nacional de Abastecimento}, Brazilian National Supply Company
(CONAB)~\citep{mendes2019calendario}.

Training a model that uses weather data as a feature has an inherent
challenge: if the weather data comes from observations, the model won't be
exposed to uncertainty when used with weather forecasts. Additionally, even
if weather forecasts are input as points, the model will not have been
exposed to uncertainty in the weather forecasts because those would be input
as point estimates. Therefore, we've introduced a regularization layer that
adds zero-centered noise to the normalized dynamic inputs of the neural
network, which doubles as a random data augmentation method.

To enable scalable crop yield estimation and risk analysis using a
data-driven model, we decided to build a model that, instead of providing
point-estimates only, can generate probability distributions. Therefore, we
decided to extend our previous works~\citep{oliveira2018scalable,
cunha2020estimating} in two different ways: (i)by  making the model more precise
by adding easy-to-obtain domain knowledge to it and (ii) altering the model
to generate (parametric) probability distributions instead of producing only
point estimates. Such a change enables us to compute the spread of the
distribution and, in doing so, assessing the uncertainty inherent in
predictions made by the model.

\subsection{On uncertainty estimates}

To explain the transition from point estimates to distributions, the key
insight needed is that parametric distributions can be represented by a small,
fixed set of parameters. Therefore, an existing model can be extended from
providing point estimates to distributions by making it predict the
\emph{parameters} of a pre-selected distribution.  Having inputs to our neural
network scaled to the line segment $[0, 1]$, and labels in the training set
also scaled to $[0, 1]$, an argument can be made to use a distribution with
support in non-negative real numbers $\mathbb{R}_{\ge
0}$\footnote{
  One might argue that only support is the line segment $[0, 1]$ is required,
  but one must be aware that record-high yields in the test set are
  a possibility. For these cases, the yield will be greater than $1$. Since we
  cannot say beforehand how much bigger yields might get, we decided to settle
  to $\mathbb{R}_{\ge 0}$.
}. While evaluating our neural network design, though, we observed that using
a normal distribution yielded better results, and we proceeded with the 
Probability Distribution Function (PDF)
\begin{equation}
  f(x; \mu, \sigma) = \frac{1}{\sigma\sqrt{2\pi}}e^{-\frac{1}{2}\left(\frac{x-\mu}{\sigma}\right)^2}\text{,}
\label{eq:normal}
\end{equation}
which means our neural network approximates parameters $\mu\in\mathbb{R}$,
$\sigma>0$. 

This leaves us with the challenge of designing a loss function such that we can
learn the parameters of the distribution by gradient descent. One such
candidate function is the negative log-likelihood function which, for the
normal distribution, has the  analytical form
\begin{equation}
  \mathrm{nll}(x_1, \ldots, x_n; \mu, \sigma) =
    \frac{n}{2}{\left(\ln(2\pi\sigma^2)\right)} +
    \frac{1}{2\sigma^2}\sum_{i=1}^{n}{(x_i-\mu)}^2\text{.}\label{eq:nll}
\end{equation}

As we're minimizing the negative log-likelihood, this is equivalent to
maximizing the log-likelihood. Moreover, given that the logarithm function is
monotonic, maximizing the log-likelihood also maximizes the
likelihood function. Therefore, the loss function we used will learn to
approximate parameters $\mu_i, \sigma_i$ of a probability distribution that will maximize the likelihood of
observed events. During the development of the model, we noticed that the model
could collapse the distribution to a small range by making $\sigma \rightarrow
0$. Due to that, we've combined equation~\eqref{eq:nll} with the entropy
of~\eqref{eq:normal}
\begin{equation}
  H(\mu, \sigma) = \frac{1}{2}\ln(2e\pi\sigma^2)\text{,}\label{eq:entropy}
\end{equation}
yielding the final loss
\begin{equation}
  l(x_1, \ldots, x_n; \mu, \sigma) = \mathrm{nll}(x_1, \ldots, x_n; \mu,
  \sigma) - \eta H(\mu, \sigma)\text{,}
\end{equation}
where $\eta$ is a hyperparameter that tells how strong is the entropy
regularization.

We also employed a Gaussian prior on the weights
\begin{equation}
	\pi(\mathbf{W}) \propto \prod_{l=1}^{L}\prod_{i,j}e^{-\frac{1}{2}(W_{i,j}^{(l)})^2}\text{,}
\end{equation}
where $L$ is the number of layers in the neural network, and $i, j$ are
indices in the weight matrices, and $\mathbf{W}$ is a weight matrix. Notice
the prior above is equivalent to a weight-decay
penalty~\citep{vladimirova2019understanding}, which is what we implemented.
Our implementation also uses dropout, which has an interpretation as
approximate Bayesian inference~\citep{gal2016dropout}. With this model
implementation we were able to improve upon the results of previous
iterations of the model, while also being able to generate uncertainty
information together with average predictions.

\subsection{Results}

We evaluate soybean and corn yield for Brazil in 2018 and compare it with
previous work. We show that not only our estimates are better than with point
estimates, but also that uncertainty information is useful to interpret the
output of the model. In this work, we use the following metrics to estimate
yield predictions quality.

\begin{enumerate}
    \item Pearson correlation coefficient (Correlation).
        \begin{equation}
            \rho_{S, O}=\frac{\operatorname{cov}(S, O)}{\sigma_{S} \sigma_{O}},
        \end{equation}
        where $\operatorname{cov}$ is the covariance, $\sigma_{S}$ is the standard deviation of $S$.
        $\sigma_{O}$ is the standard deviation of $O$. Here $S$ stands for predicted values and $O$
        observed (measured) yield values.

    \item Percentual root mean square error (PRMSE). 
    \begin{equation}
        \mathrm{PRMSE}=\frac{\sqrt{\frac{1}{n} \sum_{i=1}^{n}\left(s_{i}-o_{i}\right)^{2}}}{\bar{o}},
    \end{equation}
    where $o_i$ is the measured yield for the county, $s_{i}$ are predicted
    values of the yield, $n$ the number of observations, and $\bar{o}$ is the mean observed value.

    \item Mean absolute percentage error (MAPE).
        \begin{equation}
            \mathrm{M}=\frac{1}{n} \sum_{i=1}^{n}\left|\frac{s_{i}-o_i}{o_i}\right|
        \end{equation}
    where $O$ is the measured yield for the county, $S_{i}$ are yield predicted
    values, and $n$ the number of observations.
\end{enumerate}

We compared the results from our previous work~\citep{cunha2020estimating} for
the estimation of Soybean productivity in Brazil for 2018 and followed the same
training procedure used there with the same model setup and hyperparameters.
For the hyperparameters introduced in this paper (noise and entropy
regularization), we performed 66 trials of hyperparameter optimization with
Optuna~\citep{optuna_2019}, which yielded noise = $0.0824$, and $\eta=0.2359$
for the soybean model, and noise = $0.0984$, and $\eta=0.3451$ for the corn model.
The other hyperparameters were: learning rate $\alpha=5\times10^{-4}$,the maximum
number of epochs set to $1000$, with early stopping patience set to $50$.

\begin{table}[h]
	\centering
  \scriptsize
	\begin{tabular}{lccc}
		\toprule
		Model & Culture & Correlation & MAPE \\
		\midrule
		\citet{cunha2020estimating} & Soybean & 0.300 & 16.015 \\
		Ours & Soybean & \textbf{0.453} & \textbf{13.010} \\
		\midrule \\
		\citet{cunha2020estimating} & Corn & 0.881 & 47.855 \\
		Ours & Corn & \textbf{0.912} & \textbf{43.063} \\
		\bottomrule
	\end{tabular}
	\caption{
		Comparison of results between previous and current model
		for Soybean and Corn crops in Brazil for 2018. Values
		in bold designate better performance.
	}\label{tab:nn-results}
\end{table}

\begin{figure}[h]
  \begin{subfigure}{.5\textwidth}
  \centering
  \includegraphics[width=.8\linewidth]{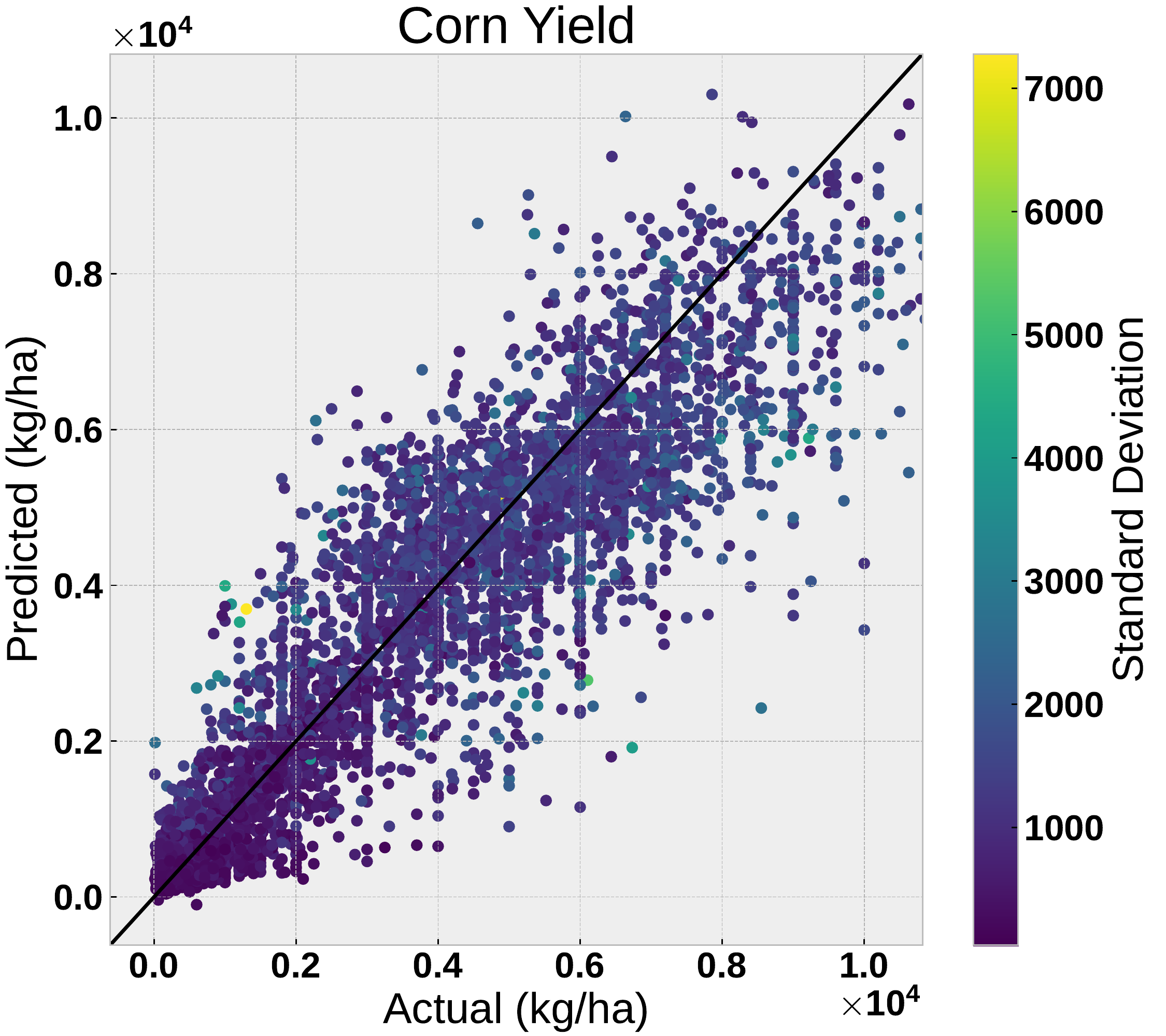}
  \caption{Yield estimates for corn (Brazil-2018).}
  \end{subfigure}
  \begin{subfigure}{.5\textwidth}
  \centering
  \includegraphics[width=.8\linewidth]{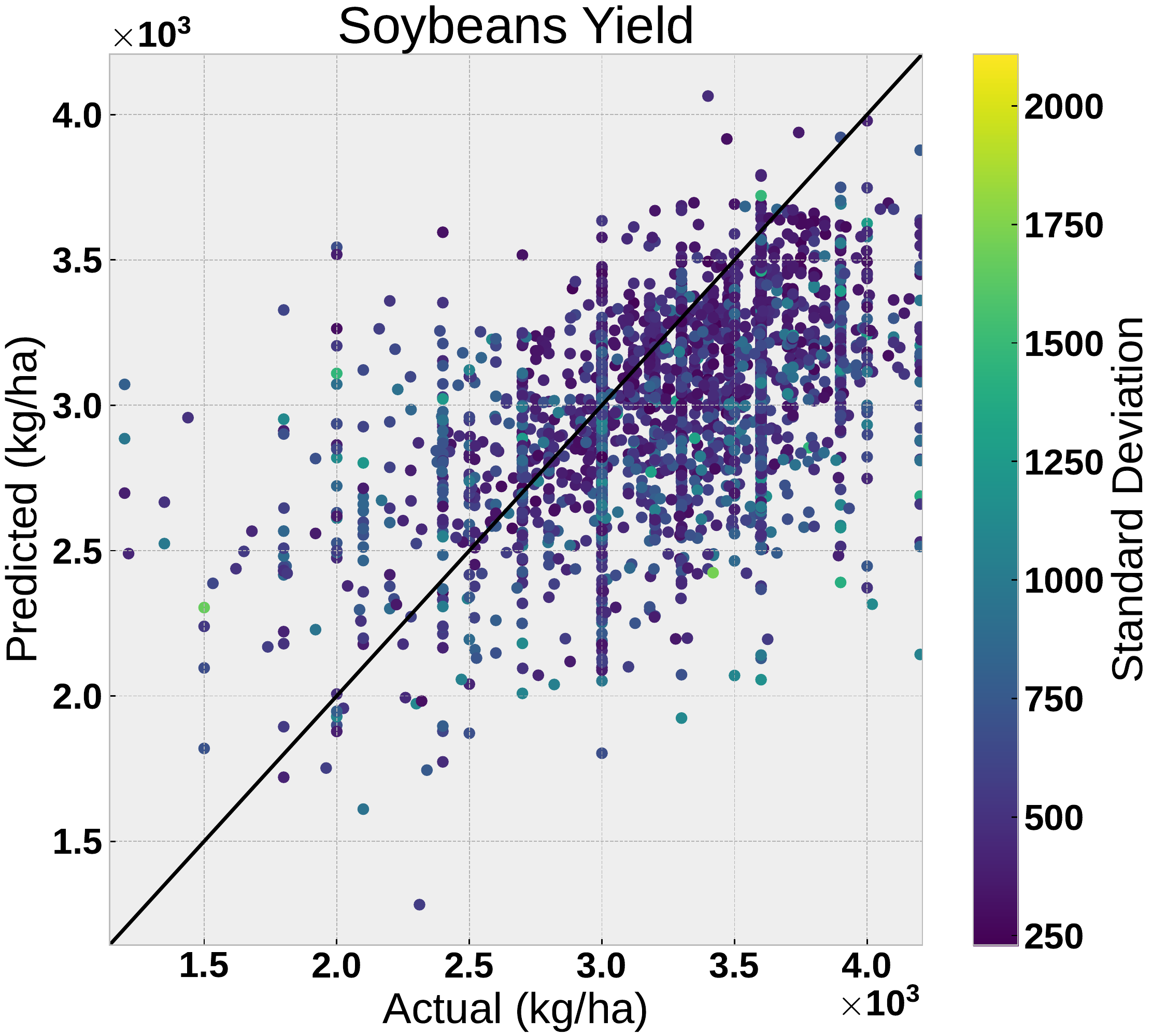}
  \caption{Yield estimates for soybeans (Brazil-2018).}
  \end{subfigure}
  \caption{%
     Predictions are improved by having an associated uncertainty level represented by standard deviation.
    As can be seen in the graph, most extreme standard deviation values are
    also the ones with higher associated error.
  }\label{fig:nn-yield}
\end{figure}

In Table~\ref{tab:nn-results}, we see a summary of the results for soybeans
and corn. As can be seen, the neural network yields more accurate results
than the previous iteration, \emph{while also providing uncertainty
information}. Uncertainty information is highlighted in
Figure~\ref{fig:nn-yield}. As implied from the graphs, higher uncertainty
predictions can have different treatment than predictions with less
uncertainty.

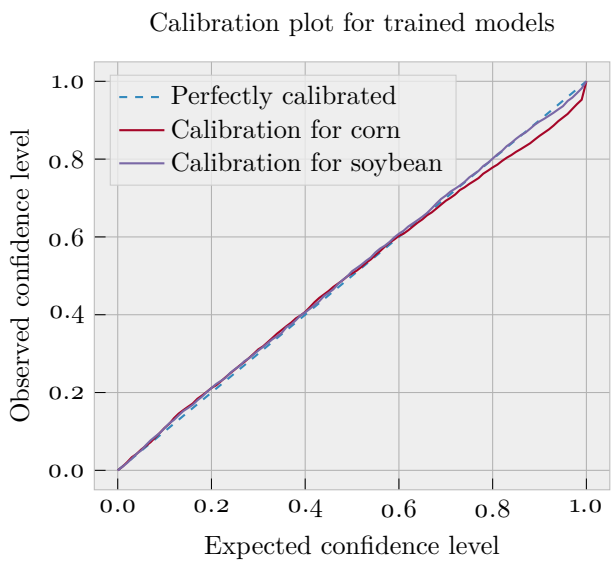
\begin{figure}
        \centering
\begin{tikzpicture}

\definecolor{color0}{rgb}{0.203921568627451,0.541176470588235,0.741176470588235}
\definecolor{color1}{rgb}{0.650980392156863,0.0235294117647059,0.156862745098039}
\definecolor{color2}{rgb}{0.47843137254902,0.407843137254902,0.650980392156863}

\begin{axis}[
axis background/.style={fill=white!93.3333333333333!black},
axis line style={white!73.7254901960784!black},
legend cell align={left},
legend style={
  fill opacity=0.8,
  draw opacity=1,
  text opacity=1,
  at={(0.03,0.97)},
  anchor=north west,
  draw=white!80!black,
  fill=white!93.3333333333333!black
},
tick pos=left,
title={Calibration plot for trained models},
x grid style={white!69.8039215686274!black},
xlabel={Expected confidence level},
xmajorgrids,
xmin=-0.05, xmax=1.05,
xtick style={color=black},
xtick={-0.2,0,0.2,0.4,0.6,0.8,1,1.2},
xticklabels={\ensuremath{-}0.2,0.0,0.2,0.4,0.6,0.8,1.0,1.2},
y grid style={white!69.8039215686274!black},
ylabel={Observed confidence level},
ymajorgrids,
ymin=-0.05, ymax=1.05,
ytick style={color=black},
ytick={-0.2,0,0.2,0.4,0.6,0.8,1,1.2},
yticklabels={\ensuremath{-}0.2,0.0,0.2,0.4,0.6,0.8,1.0,1.2}
]
\addplot [thick, color0, dashed]
table {%
0 0
0.01 0.01
0.02 0.02
0.03 0.03
0.04 0.04
0.05 0.05
0.06 0.06
0.07 0.07
0.08 0.08
0.09 0.09
0.1 0.1
0.11 0.11
0.12 0.12
0.13 0.13
0.14 0.14
0.15 0.15
0.16 0.16
0.17 0.17
0.18 0.18
0.19 0.19
0.2 0.2
0.21 0.21
0.22 0.22
0.23 0.23
0.24 0.24
0.25 0.25
0.26 0.26
0.27 0.27
0.28 0.28
0.29 0.29
0.3 0.3
0.31 0.31
0.32 0.32
0.33 0.33
0.34 0.34
0.35 0.35
0.36 0.36
0.37 0.37
0.38 0.38
0.39 0.39
0.4 0.4
0.41 0.41
0.42 0.42
0.43 0.43
0.44 0.44
0.45 0.45
0.46 0.46
0.47 0.47
0.48 0.48
0.49 0.49
0.5 0.5
0.51 0.51
0.52 0.52
0.53 0.53
0.54 0.54
0.55 0.55
0.56 0.56
0.57 0.57
0.58 0.58
0.59 0.59
0.6 0.6
0.61 0.61
0.62 0.62
0.63 0.63
0.64 0.64
0.65 0.65
0.66 0.66
0.67 0.67
0.68 0.68
0.69 0.69
0.7 0.7
0.71 0.71
0.72 0.72
0.73 0.73
0.74 0.74
0.75 0.75
0.76 0.76
0.77 0.77
0.78 0.78
0.79 0.79
0.8 0.8
0.81 0.81
0.82 0.82
0.83 0.83
0.84 0.84
0.85 0.85
0.86 0.86
0.87 0.87
0.88 0.88
0.89 0.89
0.9 0.9
0.91 0.91
0.92 0.92
0.93 0.93
0.94 0.94
0.95 0.95
0.96 0.96
0.97 0.97
0.98 0.98
0.99 0.99
1 1
};
\addlegendentry{Perfectly calibrated}
\addplot [thick, color1]
table {%
0 0
0.01 0.00929956470122675
0.02 0.0207756232686981
0.03 0.0336367233874159
0.04 0.0419469726948951
0.05 0.0534230312623664
0.06 0.0629204590423427
0.07 0.0722200237435695
0.08 0.0842896715472893
0.09 0.0965571824297586
0.1 0.109220419469727
0.11 0.119509299564701
0.12 0.134151167392165
0.13 0.145825089038385
0.14 0.154926790660863
0.15 0.163434903047091
0.16 0.171745152354571
0.17 0.18401266323704
0.18 0.192520775623269
0.19 0.201820340324495
0.2 0.211713494261971
0.21 0.218638702018203
0.22 0.228729719034428
0.23 0.239216462208152
0.24 0.249505342303126
0.25 0.259200633161852
0.26 0.269291650178077
0.27 0.278195488721804
0.28 0.288682231895528
0.29 0.299564701226751
0.3 0.311040759794222
0.31 0.318757419865453
0.32 0.329442026117926
0.33 0.340324495449149
0.34 0.351404827859121
0.35 0.360902255639098
0.36 0.370597546497824
0.37 0.379699248120301
0.38 0.390581717451524
0.39 0.398100514444005
0.4 0.407202216066482
0.41 0.419469726948951
0.42 0.432330827067669
0.43 0.443213296398892
0.44 0.45211713494262
0.45 0.461020973486347
0.46 0.471507716660071
0.47 0.4800158290463
0.48 0.488721804511278
0.49 0.498812821527503
0.5 0.506133755441235
0.51 0.513454689354966
0.52 0.522754254056193
0.53 0.533438860308666
0.54 0.542144835773645
0.55 0.553620894341116
0.56 0.563711911357341
0.57 0.574990106846063
0.58 0.585081123862287
0.59 0.593391373169767
0.6 0.602690937870993
0.61 0.608428967154729
0.62 0.617134942619707
0.63 0.627028096557182
0.64 0.636723387415908
0.65 0.645033636723387
0.66 0.655520379897111
0.67 0.663039176889592
0.68 0.672932330827068
0.69 0.683023347843292
0.7 0.692916501780768
0.71 0.700435298773249
0.72 0.708943411159478
0.73 0.720419469726949
0.74 0.727344677483182
0.75 0.73605065294816
0.76 0.743569449940641
0.77 0.751286110011872
0.78 0.762564305500594
0.79 0.770280965571824
0.8 0.778591214879303
0.81 0.785714285714286
0.82 0.795211713494262
0.83 0.803521962801741
0.84 0.810447170557974
0.85 0.818757419865453
0.86 0.825682627621686
0.87 0.833201424614167
0.88 0.840720221606648
0.89 0.850415512465374
0.9 0.859715077166601
0.91 0.86802532647408
0.92 0.876335575781559
0.93 0.886228729719034
0.94 0.897506925207756
0.95 0.906806489908983
0.96 0.91610605461021
0.97 0.928967154728928
0.98 0.941432528690146
0.99 0.952710724178868
1 1
};
\addlegendentry{Calibration for corn}
\addplot [thick, color2]
table {%
0 0
0.01 0.00907127429805616
0.02 0.0190064794816415
0.03 0.0302375809935205
0.04 0.0423326133909287
0.05 0.0509719222462203
0.06 0.0647948164146868
0.07 0.076889848812095
0.08 0.0868250539956803
0.09 0.0963282937365011
0.1 0.108423326133909
0.11 0.120950323974082
0.12 0.129589632829374
0.13 0.141252699784017
0.14 0.152483801295896
0.15 0.160259179265659
0.16 0.168466522678186
0.17 0.180129589632829
0.18 0.18963282937365
0.19 0.201295896328294
0.2 0.211663066954644
0.21 0.222030237580994
0.22 0.229805615550756
0.23 0.237580993520518
0.24 0.250539956803456
0.25 0.260043196544276
0.26 0.268250539956803
0.27 0.280345572354212
0.28 0.288120950323974
0.29 0.298056155507559
0.3 0.307991360691145
0.31 0.317062634989201
0.32 0.326997840172786
0.33 0.336933045356371
0.34 0.344708423326134
0.35 0.353347732181426
0.36 0.368034557235421
0.37 0.374946004319654
0.38 0.385745140388769
0.39 0.398272138228942
0.4 0.406047516198704
0.41 0.41511879049676
0.42 0.425053995680346
0.43 0.434557235421166
0.44 0.447948164146868
0.45 0.456155507559395
0.46 0.465226781857451
0.47 0.478617710583153
0.48 0.489416846652268
0.49 0.498920086393089
0.5 0.511879049676026
0.51 0.520086393088553
0.52 0.528725701943844
0.53 0.53866090712743
0.54 0.547300215982721
0.55 0.560259179265659
0.56 0.570194384449244
0.57 0.577537796976242
0.58 0.587041036717063
0.59 0.598704103671706
0.6 0.608207343412527
0.61 0.615982721382289
0.62 0.626781857451404
0.63 0.634989200863931
0.64 0.643196544276458
0.65 0.651835853131749
0.66 0.66133909287257
0.67 0.673434125269978
0.68 0.685961123110151
0.69 0.697624190064795
0.7 0.706263498920086
0.71 0.716630669546436
0.72 0.723974082073434
0.73 0.732181425485961
0.74 0.741252699784017
0.75 0.752915766738661
0.76 0.761555075593953
0.77 0.769762419006479
0.78 0.782289416846652
0.79 0.790496760259179
0.8 0.801295896328294
0.81 0.812095032397408
0.82 0.821166306695464
0.83 0.833261339092873
0.84 0.841900647948164
0.85 0.852267818574514
0.86 0.86133909287257
0.87 0.868250539956803
0.88 0.878185745140389
0.89 0.888552915766739
0.9 0.896760259179266
0.91 0.903671706263499
0.92 0.911447084233261
0.93 0.919222462203024
0.94 0.927861771058315
0.95 0.936069114470842
0.96 0.949460043196544
0.97 0.958099352051836
0.98 0.96976241900648
0.99 0.981857451403888
1 1
};
\addlegendentry{Calibration for soybean}
\end{axis}

\end{tikzpicture}
        \caption{%
                Calibration~\citep{kuleshov2018accurate} plots for both trained models.
                Both models were trained and evaluated independently, with new
                noise and entropy hyperparameters optimized independently, but sharing
                others, such as number of epochs, learning rate, and early stopping
                patience.
        }\label{fig:calibration}
\end{figure}

To assess whether the uncertainty estimates of the models are useful, we also
plotted calibration diagrams~\citep{kuleshov2018accurate}\footnote{
	Not to be confused with the calibration process of simulation models,
	presented in the next section.
} for both models.
Calibration means that a predicted value $y_i$ should fall in a X\%
confidence interval X\% of the time. Formally, when $x_i, y_i$ are independent,
identically-distributed realizations of random variables $X, Y$, calibration
can be computed by
\begin{equation*}
	P(Y\le F_X^{-1}(p)) = p \text{ for all } p \in [0, 1]\text{,}
\end{equation*}
where $F_X$ is the forecast value at $X$. The calibration plots are shown in
Figure~\ref{fig:calibration}, and we can see that both trained models are close
to the ideal identity line, although the confidence expressed by the corn model
is smaller than it could have been, especially for larger confidence levels.

\section{Crop Simulation Model}\label{sec:crop-simulation-model}

A key challenge when dealing with crop yield forecasting is the lack of
field-level resolution ground truth data globally available. This fact makes
data-driven crop models that perform well for a given location not useful for
estimating yield in other places. In other words, a new data-driven model
should be created for each new evaluation crop/location. The overcome this
issue, we proposed the utilization of crop simulation models (e.g., DSSAT
\citep{hoogenboom2019dssat}) that can be calibrated to support the yield
prediction in different areas. Crop simulation models are based on a detailed
representation of plant physiology and the biological response to weather,
soil, genetic and management inputs. There are several metrics that can be
estimated with the use of crop simulation models such as yield, above-ground
biomass, leaf area index, physiological maturity, and seed density. Another
important feature of crop simulation models is the ability of computing
output metrics during the season.

We created a crop simulation wrapper (see Figure~\ref{fig:dssat_wrapper}) to
support the utilization of multiple crop models in a transparent way. So we
can provide the input data and expect the same output for different crop
simulation models. The provided data is composed of management, weather,
genetics, and soil data. We employ JSON data format for passing and receiving
the information from a higher-level application to the crop simulation wrapper. The
data input is as follows:

\begin{figure}[h]
  \centering
  \includegraphics[width=\linewidth]{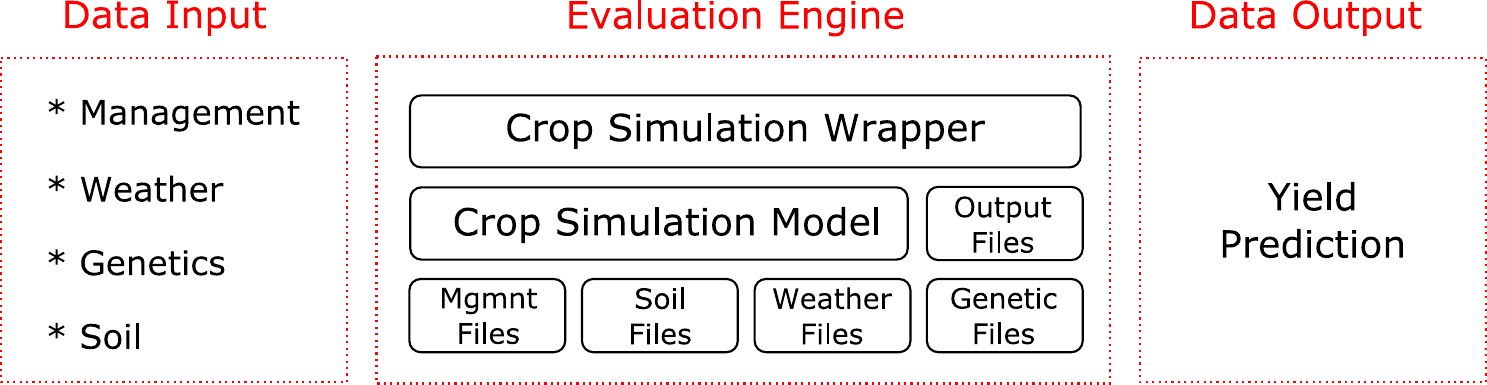}
  \caption{%
    Crop simulation wrapper. A transparent data representation format is
used to support different crop simulation models. Input data
comprises: (i) management, (ii) weather, (iii) genetics, and (iv) soil
properties.  Output data includes leaf area index, evapotranspiration, and
crop yield.
  }\label{fig:dssat_wrapper}
\end{figure}

\begin{enumerate}

\item \textbf{Management Data.}  The management data includes planting and
harvest dates, the description of initial soil conditions, such as information
about the previous crop, and fertilization and irrigation dates and methods.

\item \textbf{Weather Data.} The weather data should be provided in a daily
format and must include temperature maximum/minimum, solar irradiation,
precipitation, and wind speed. We created a service to globally collect data
from different data sources (including IBM Pairs) and provide it in a format
compatible with the proposed crop simulation wrapper
(Figure~\ref{fig:dssat_wrapper}).

\item \textbf{Genetics Data.} The Genetics characteristics refer to plant
coefficients related to the development, growth, and yield among different
cultivars when planted in the same environment. Such genetic coefficients can
be vegetative or reproductive. In the case of soybean, for instance, there are
parameters for representing the average time between the first flower and the
first seed. There are also parameters for estimating the maximum size of a full
leaf and the mean fraction of oil/protein in the seeds.

The genetic coefficients can be measured directly on the field or estimated
using inversion techniques~\citep{sun2015model}. For instance, we could use the measured yield
to assess the genetic coefficients that best fit the crop model output. We
created an inversion engine to estimate genetic parameters for a given area
(Section~\ref{sec:calibration}). 

\item \textbf{Soil Data.} We obtain soil data from
SoilGrids~\citep{hengl2017soilgrids250m} dataset, which has soil information
(both observed and model-generated data) in a 250m grid for the whole planet.
It provides seven layers of soil data, in which we used  nine features: clay
content, silt content, sand content, bulk density, coarse fragments, cation
exchange capacity, organic carbon content, pH in $H_2O$, and pH in
$K\mathrm{Cl}$. We also created a service for converting SoilGrids data into the
crop simulation model format.

\end{enumerate}

\subsection{Crop Model Calibration Service}\label{sec:calibration}

The proposed framework calibrates the crop simulation model for each
combination of crop/location to determine the genetic coefficients. Instead of
just using direct measurements on the field, we use known input parameters
(e.g., soil, and weather) and past yield data to estimate unknown parameters.
To do this task, we employ the derivative-free optimization method Particle
Swarm Optimization (PSO) \citep{audet2017derivative}.

\begin{figure}[h]
  \centering
  \includegraphics[width=\linewidth]{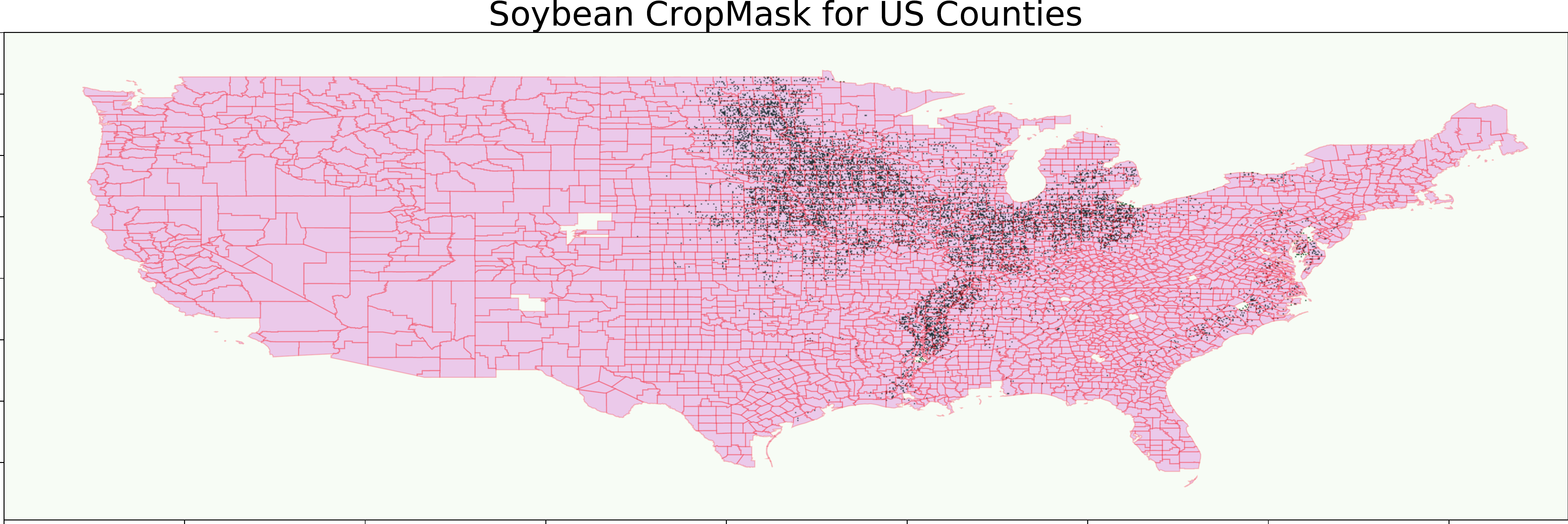}
  \caption{%
      CropScape mask for soybean in US (2018).
  }\label{fig:cmask}
\end{figure}

\begin{figure}[h]
  \centering
  \includegraphics[width= 0.7\linewidth]{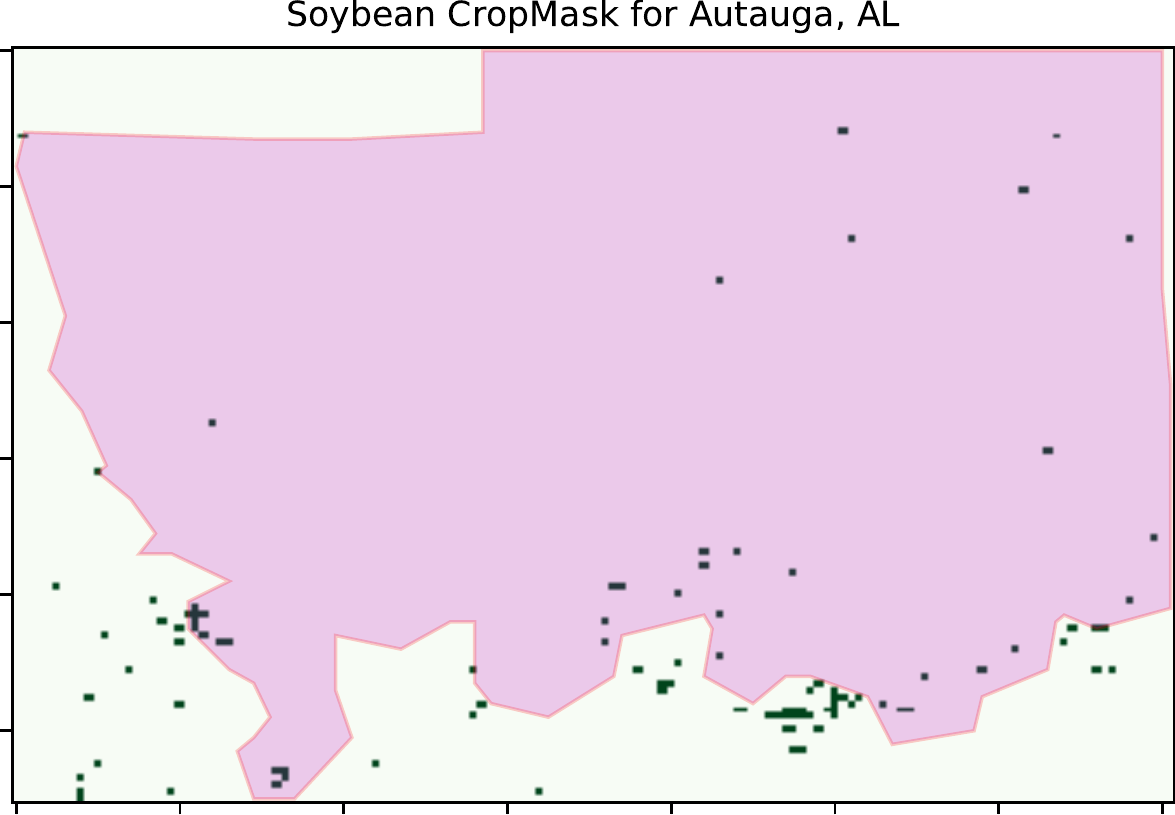}
  \caption{%
      CropScape mask for soybean in Autauga, AL (2018).
  }\label{fig:ccmask}
\end{figure}

We created a system and method to process multiple data sources and evaluate
different combinations of year, latitude, longitude, yield, state, and
county. To use the calibration service, the user should provide a map with
the vector representation of different regions and where the farms are
located. For instance, we employed the CropScape
environment~\citep{han2012cropscape} which provides conterminous geospatial
cropland data for the entire United States. By using this service, we can
estimate where and when a given crop was planted. Figure~\ref{fig:cmask}
shows the location of soybean fields in the US in 2018 and
Figure~\ref{fig:ccmask} presents the crop mask for Autauga, AL in 2018. The
black points show the soybean fields for the selected region. Quick Stats
service \citep{nass2011quick} delivers county yield data for a given year. As
we have multiple fields in a given county, we assume they had the same yield
per year. Table \ref{table:cservinput} presents a sample of input data for
crop model calibration service. As we may have a relatively large number of
fields to be calibrated, we created a service to parallelize the calibration
process for all the fields. The service is on the top of the Celery
framework\footnote{https://docs.celeryproject.org}, which is a distributed
python framework to communicate different services.

\begin{figure*}[!h]
  \centering
  \includegraphics[width=0.75\linewidth]{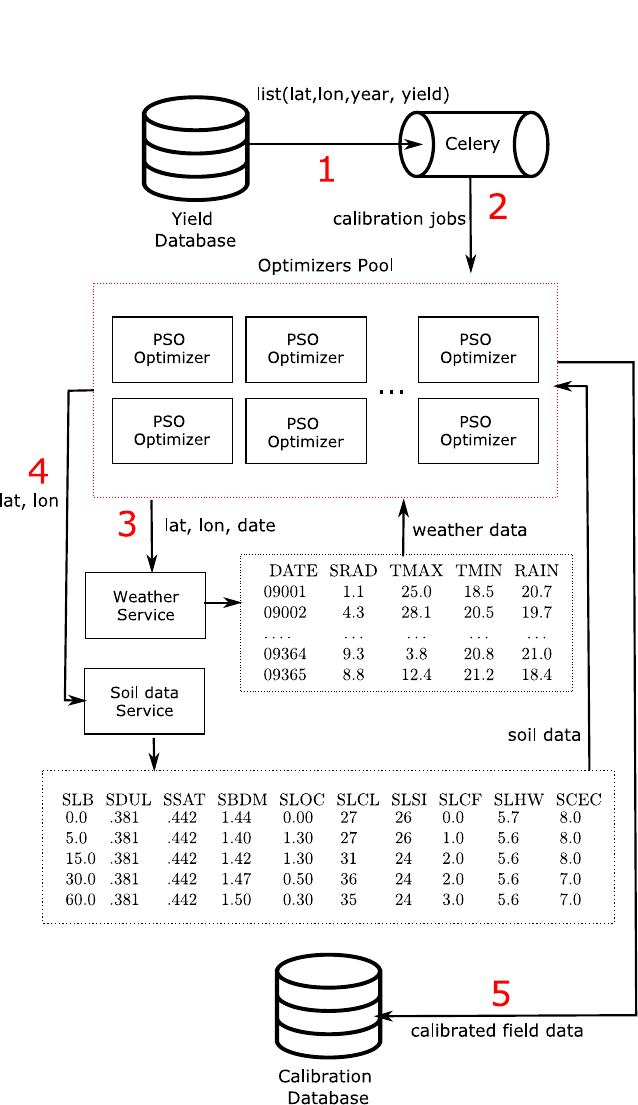}
  \caption{
    Calibration service. It receives the past yield data as input (1), creates a
    job for each tuple, and sends them via celery to the optimizers pool (2).
    Each optimizer obtains data from weather and soil services (3) and (4) to
    execute the genetic coefficients inversion. Finally, the calibration
    service sends the calibrated data for each field to the calibration
    database (5). 
  }\label{fig:calibrationservice}
\end{figure*}

\begin{table}
	\centering
  \scriptsize
    \begin{tabular}{ccccccc}
    \toprule
     year &  lat & lon & fips & yield & state & county \\
    \midrule
     2013 &  40.106 & -90.000 & 17017 & 3537 & IL & Cass \\
     2013 &  40.099 & -90.003 & 17017 & 3537 & IL & Cass \\
     2013 &  40.095 & -90.018 & 17017 & 3537 & IL & Cass \\
     2013 &  40.088 & -90.061 & 17017 & 3537 & IL & Cass \\
     2013 &  40.088 & -90.054 & 17017 & 3537 & IL & Cass \\
     2013 &  40.088 & -90.050 & 17017 & 3537 & IL & Cass \\
    \bottomrule
    \end{tabular}
\caption{
    Example of aggregated data for crop model calibration service stored in
    yield databse. The full dataset used in this paper has $4114524$ rows and
    yield data from 2010 to 2018.
}
\label{table:cservinput}
\end{table}

Figure~\ref{fig:calibrationservice} depicts an overview of how the calibration
service obtains the yield data (e.g., Table~\ref{table:cservinput}) and ends up
with the genetic coefficients for each field using inversion techniques such as
PSO.  The calibration process executes in parallel by using data provided by
the yield database (1). Each calibration job receives a field entry with
latitude, longitude, year, and the corresponding yield in that year (2).
Weather and soil data services (3) and (4) collect and process data from the
respective data sources and pass the formatted data to the optimization job.
Finally, the calibration job executes the PSO optimization to assess what are
the genetic coefficients, planting, and harvest dates that provide the closest
yield compared to the passes in step (1). The cost metric ($C$) is calculated
as follows:

\begin{equation}
    C = \alpha \times \frac{|Y_{sim} - Y_{meas}|}{Y_{meas}} + \beta \times \frac{\text{RMSE}_{LAI}}{\mu_{LAI}},
\end{equation}

where $\alpha$ corresponds to the yield importance hyperparameter, $Y_{sim}$ is
the simulated yield, $Y_{meas}$ represents the measured yield, $\beta$
represents the LAI importance hyperparameter, $\text{RMSE}_{LAI}$ is the root
mean squared error for leaf area index of simulated and the estimated leaf area
index for a given area, and  $\mu_{LAI}$ corresponds to the mean value for the
leaf area index.  We evaluated the results of the calibration process with and
without using the leaf area index in the cost calculation. We realized that
using LAI in the costs equation let the yield prediction worst when compared to
the results using the cost without it (i.e., $\beta = 0$). Other works reached
the same conclusion~\citep{nearing2012}, it was realized that using LAI to
improve yield estimates was not helpful due to several reasons including
low spatial resolution and LAI estimation uncertainties.

\begin{lstlisting}[language=Python, caption=Calibrated field example, label=lst:cal,
  basicstyle=\footnotesize]
{
    "Adams": {
        "2012": [{
            "calibration_values": [
                0.1574, 0.6511,
                0.1693, 0.1119,
                ...
                0.3883
            ],
            "calibration_cost": 0.0617,
            "location": {
                "latitude": 39.8428,
                "longitude": -91.2100,
                "measured_yield": 2737.0
            }
        }]
    }
}
\end{lstlisting}

Listing \ref{lst:cal} presents a sample calibration result. For each county
(e.g., Adams), we have a set of calibrated years (e.g., 2012).  The calibration
service creates a list of calibrated fields where each item has the following
information: evaluated coefficients (including plant/harvest dates), optimization cost,
latitude, longitude, and the measured yield in that year.

\subsection{Crop Model Evalution Service}\label{sec:evaluation}

We created a similar service to the calibration one on top of a
high-performance computing (HPC) infrastructure to predict crop yield for
large areas (Figure~\ref{fig:calibrationservice}). The Crop Model Evaluation
Service has a pool of evaluators instead of PSO optimizers, the calibration
database is the service input, and the output of this service is the
prediction database. Another difference is the PSO optimizers take several
simulations to find the best fit for the genetic coefficients. In the
evaluation service, each evaluator in the pool runs the crop simulation model
just a single time for each field (pixel). In this case, a region (e.g.,
county) may have multiple calibrated fields but only a single target yield
during calibration as we have (in most cases) just yearly measured yield.

The calibration database stores genetic coefficients for each field
considering different years. Then, when estimating the yield for a given farm
in a new year, what calibrated coefficients should we use? There are
different forms to use calibrated models from previous years for predicting
yield on a given farm. For instance, we can use the calibrated model from the
last year to estimate field productivity. However, this approach disregard
calibrated models from previous years that can better predict productivity
for the current year. In this section, we present different forms to group
calibrated models to estimate yield for a given farm.

Assume we are interested in estimating crop yield productivity for a given
year $y_i$ and we have a set of $n$ calibrated models ${M(y_i-n), M(y_i-n+1),
\cdots, M(y_i-1))}$ where $M(y_x)$ denotes a model calibrated using data
collected in year $y_x$. The predicted yield for a year $y_i$ using a
calibrated model $M(y_x)$ where $y_x < y_i$ is denoted by $Y_d(y_x, y_i)$. We
can use the following approaches to estimate yield $Y_d(y_i)$ a single field
using a combination of previously calibrated models. 

\begin{itemize}
  
  \item \textbf{All previous calibrations}. This ensemble method is the
  simplest one and uses a random calibrated model to predict the yield
  (Equation \ref{eq:previous_cal}). For instance, to evaluate the yield for
  McLean county in 2012, the crop model evaluation service executes the model
  with calibrated parameters from 2011, 2010, 2009, and 2008. The user can
  pick one to perform her evaluation.

  \begin{equation}
  \label{eq:previous_cal}
      Y_d(y_i) = Y_d(y_x, y_i), y_x < y_i 
  \end{equation}

  \item \textbf{Previous calibrated year}. This approach uses the last calibrated year to 
  predict the crop yield (Equation \ref{eq:last_cal}).

  \begin{equation}
  \label{eq:last_cal}
      Y_d(y_i) = Y_d(y_{i - 1}, y_i)
  \end{equation}

  \item \textbf{Mean of previous calibrations}. Employs the mean value of $n$
  previously calibrated models to estimate crop productivity (Equation
  \ref{eq:mean_cal}).

  \begin{equation}
  \label{eq:mean_cal}
    Y_d(y_i) = \sum_{j=y_i-n}^{y_i-1}\frac{Y_d(y_j, y_i)}{n}
  \end{equation}

  \item \textbf{Quality-based ensemble}. We used a weighted average approach
  using the MAPE metric to determine the weight of each ensemble component
  (Equation \ref{eq:quality_ensemble}). To estimate the quality of a
  previously calibrated model, we use the mean MAPE value of all possible
  future predictions for this model. 
  To evaluate the crop yield using the quality-based ensemble with $(n-1)$
  calibrated models we use the following equation:

  \begin{equation}
  \label{eq:quality_ensemble}
    Y_d(y_i) = \frac{\sum_{j=y_i-n}^{y_i-2}Y_d(y_j, y_i) \times Q(y_j)}{\sum_{j=y_i-n}^{y_i-2} Q(y_j)}.
  \end{equation}

  $Q(y_j)$ is the mean MAPE for all possible predictions $Y_d(y_j, y_k)$ where $j < k < i$. 
  For instance, suppose we evaluate the
  yield for 2012 using previous models from 2008, 2009, and 2010. If 2009 and
  2010 have similar MAPE results, and 2010 is half of the value. Then, the
  ensemble will give doubled importance to 2011 when compared to the others.

\end{itemize}

In this experiment, we study how soybean yield can be predicted using the
proposed crop simulation model service for Illinois (USA) counties in the
2010 to 2018 period. This section presents county-level
soybean predictions, however, the proposed approach is general enough to
calibrate and evaluate other crops data at the farms level. For this study, we
selected 50 pixels at random using CropScape~\citep{han2012cropscape} data to
identify soybean fields. We also employed the Quick Stats~\citep{nass2011quick}
service to obtain the measured yield for each county in a given year.
We evaluated the different ways of calibrating models ensemble presented in
Section~\ref{sec:evaluation}.

\begin{figure*}[t]
    \centering
    \includegraphics[width=0.9\linewidth]{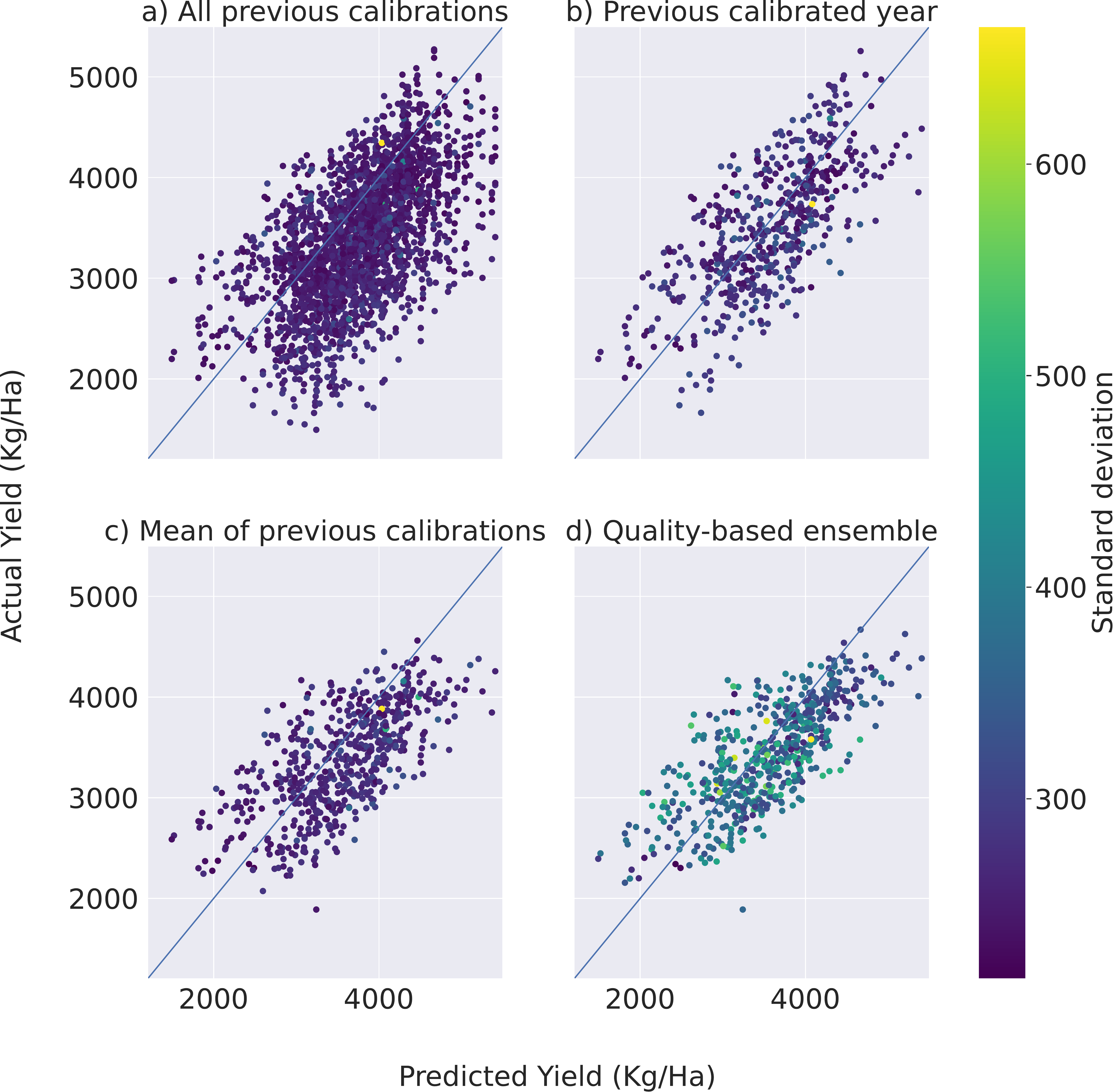}
    \caption{
        Crop simulation model results using different calibration ensembles. 
    }\label{fig:csm_results}
\end{figure*}

\begin{table}[h]
	\centering
  \scriptsize
\begin{tabular}{cccc}
\toprule
                    Experiment &  Correlation &      MAPE &     RMSEP \\
\midrule
        All previous calibrations &  0.580 &  0.146 &  0.180 \\
        Previous calibrated year &   0.705 &  0.125 &  0.144 \\
    Mean of previous calibrations &  0.669 &  0.132 &  0.150 \\
        Quality-based ensemble &     0.747 &  0.113 &  0.130 \\
\bottomrule
\end{tabular}
\caption{
    Performance metrics for different calibration ensembles.
}
\label{table:csv_results1}
\end{table}

Figure~\ref{fig:csm_results} and Table~\ref{table:csv_results1} show the
results for different ways of using calibrated models using previous seasons
data. Figure~\ref{fig:csm_results}a) shows the yield estimation results for all
possible combinations of calibrated model and further evaluation in later
years. This utilization of the previously calibrated model is pretty simple and
provides less accurate results as it does not leverage multiple calibrated
models when performing crop yield estimation. For each evaluation, we present
the mean value as the predicted value. The system calculates the standard
deviation using all the predictions for the 50 pixels calibrated within the
county for a single year.

Figure~\ref{fig:csm_results}b) depicts the results when we compare the
measured yield and the mean value of the previous calibrated models. Observe
that using the model calibrated in the last year improves the ensemble
performance. The results suggest a strong correlation between the close
planting years. One can explain this fact as farming management procedures,
and even climate change effects usually take time to change.
This hypothesis gets stronger when we observe the results considering the
mean value of previously calibrated years (Figure~\ref{fig:csm_results}c)).
In this case, the contribution to the final yield prediction of far
calibrated models is the same as close calibrated years. Observe the results,
in this case, are poor when compared to the previous calibrated year but
better than the simplest approach.

Although the previous approaches present quite good results, they can be
improved by using the other strategies to reduce the evaluation error. For
instance, as long as we use the proposed system during different years, we
can observe which calibrated model presents better predictions when compared
to others in different years. We can use a quality metric (MAPE) using
previous evaluation years to estimate how well a calibrated model tends to
perform and give more importance to this model (quality-based ensemble
strategy). Figure~\ref{fig:csm_results}d) shows the better prediction
results we found in this study as it gives more weight to calibrated models
that perform well previously. Observe this ensemble generates a higher 
standard deviation as it attributes different weights to previous calibrated
models increasing prediction variability for each pixel.

\begin{figure}[h]
  \centering
  \includegraphics[width= .99\linewidth]{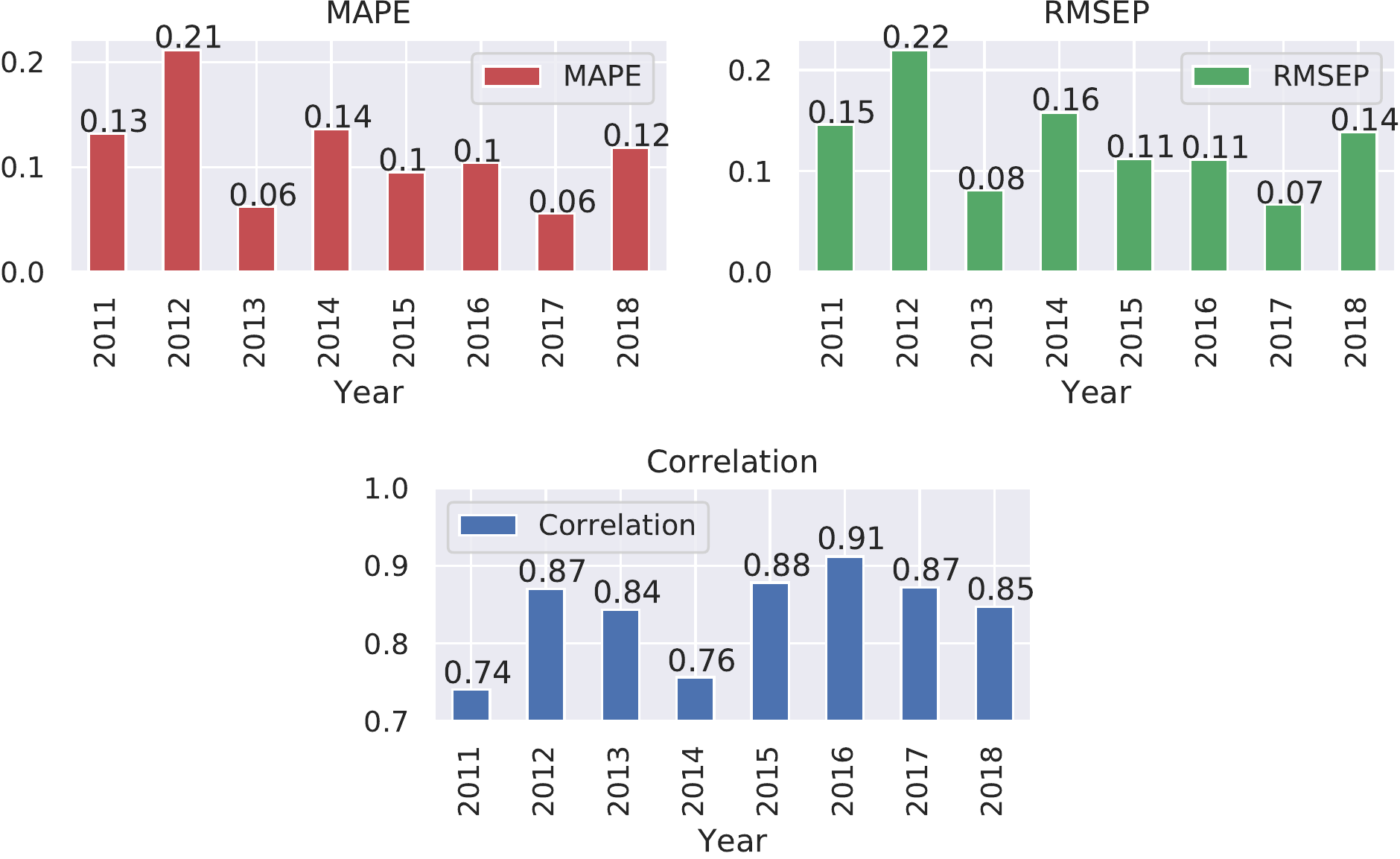}
  \caption{%
  Error-related metrics for the quality-based ensemble for calibrated for
  different years. Observe the model performance tends to improve as the
  number of years increases.
  }\label{fig:erros_time}
\end{figure}

The results that are shown in Figure~\ref{fig:csm_results}d) consider all the
evaluation years in the study interval. However, crop yield evaluation years
should have different numbers of calibration model sizes to create their
ensembles. For instance, 2012 has more calibrated models to create its
ensemble than all the others. Figure~\ref{fig:erros_time} shows how evaluation
errors change according to the number of previous calibrated years, observe 
there is a performance increase pattern when there is a higher number of previously
calibrated models. As long as there is more previous data to generate the crop simulation ensemble, 
the error tends to decrease and the correlation increases.

\section{Crop Simulation Model Surrogate}\label{sec:surrogate}

Executing a crop simulation model is computationally expensive and time-consuming.
When this task is performed a few times, just to compare, for instance,
 yield predictions under a very restricted 
set of different conditions, crop simulation models can be executed on ordinary computers. 
However, for several tasks that rely on intense sampling, such as inverse problem calculation,
risk analysis and calibration of parameters, the time spent to generate all the samples become
prohibitive for such computationally demanding models. High Performance Computing can
undoubtelly reduce the required computational time, but its usage is conditioned to access
to a very specific infrastructure that is not universally available. Essentially, to overcome 
situations in which a large number of samples from computationally demanding models are required, 
there are basically three major approaches: (i) using efficient sampling strategies, in order to 
intelligently explore the search space, extracting more information with fewer
samples; (ii) reduce the dimensionality of the search space (which means, reducing the number of input 
parameters of a model) and consequently the number of required samples; (iii) constructing surrogates
 models, that are computationally
cheaper, thus reducing the time required to perform the
sampling~\citep{frangos2010surrogate}. These strategies are not mutually exclusive and can be
used in combination with each other. In fact, by reducing the number of input parameters (approach ii), 
one must construct a surrogate model (approach iii) in order to extract the relation between
the remaining input parameters and the output of the system.

In this paper, crop simulation models were used extensively. Depending on the area to be 
evaluated the yield prediction could
take hours or even days to finish. Therefore, we decided to develop a
surrogate model for crop simulation models, as execution time was clearly a limiting condition. 
To date, there are several ways
to construct such surrogate models, and each one of them is more suitable for
some 
properties the model might present (non-linearities, number of parameters, etc). Techniques used to construct
surrogate models can
vary from simple linear regression to more sophisticated approaches, such as
Gaussian Processes (GP)~\citep{jones2009design} and Neural Networks
(NN)~\citep{tripathy2018deep}. Crop simulation models have intrinsic non-linearities and a
plurality of parameters that should be defined prior to their execution.
Therefore, naive solutions to construct surrogate models should be avoided.
More robust techniques for the construction of surrogate models are also more
computationally costly: they require the execution of the target model
several times, but this process is executed only once, offline. In this
scenario, GP and Deep Learning NN pose themselves as possible candidates for the
construction of a surrogate model. GP, however, can have a very slow
convergence and demand several (possibly unknown) assumptions for its
execution. For this reason, we opted for adopting the NN approach, which is detailed
in subsection~\ref{surrogateNN}.

The process of constructing a surrogate model consists of collecting a set of
samples that correlate different combinations of input parameters and the output
of the target system\footnote{The target system here is the expensive system
whose behaviour we are trying to emulate}. Then, the chosen technique for the
construction of the surrogate model is applied over those selected points. It is
clear that the accuracy of the surrogate model is intrinsically linked to the
set of samples used as a substrate for the inferred correlations. When selecting
the points it is important to guarantee that we have a comprehensive combination
of input parameters that cover sactisfactorily the parameters space, yet with the
fewest number of samples that guarantees a good coverage (remember that
generating a sample is time consuming). That is why we also made use of an
intelligent sampling strategy, detailed in subsection~\ref{sampling}.

We opted for not performing a model order reduction (i.e., reducing the number
of required input variables), even though crop simulation models require a large
number of them because they are mostly subject to evaluation and
optimization from crop management decision-makers. 

%
%
%
%
\subsection{Sampling strategy}
\label{sampling}

In this study, we opted to use DSSAT as the crop simulation model to conduct our
experiments. However, our strategy is generic enough to be used with others
systems. To construct the training dataset of the surrogate model, given the
extensive amount of input parameters, we opted to leave some of them fixed,
while varying properly as many as 33 parameters. We have fixed the location
(latitude and longitude), and, consequently, the parameters that describe
soil properties. We have focused on the genetic parameters of one cultivar of
soybean (totaling 18 parameters) and weather parameters (totaling 15
parameters).

Quasi-random sampling based on a Sobol sequence was the method chosen to perform
data sampling over the manifold defined by lower and upper values of the 33
variables. It is classified as a quasi-random sampling method because it first 
selects a random sample from the search space and then follows a mathematical
sequence to alter the values of the input parameters used on the subsequent
samples~\citep{kucherenko2015exploring}. Therefore, this sampling strategy is
more space-filling (i.e., it presents a good coverage of the search space,
minimizing the possibilities of concentrating samples on small subspaces). This
property is especially desirable while constructing surrogate models. This
approach outperforms pure random Monte Carlo
sampling~\citep{saltelli2010variance}.

With this strategy, we were able to generate the values for the crop management
part of the input data. For the weather parameters, DSSAT requires daily data. So,
we collected soil data from SoilGrids and historical weather data from ERA\-5 for this fixed
latitude/longitude point. We have calculated the minimum and maximum average
temperature for the 6 months that represent the seeding, growing and planting
period of soybean. The quasi-random sampling generated samples from weather parameters 
with a range of variation between $\pm$ 0.15 * average value of the parameter. 
However, DSSAT requires data on a daily granularity. Hence, to generate coherent values
on a daily bases we used the value sampled with the quasi-random method as the mean
of a normal distribution. Then, for each day of a given month, we generated a value, 
randomly sampled from this normal distribution. 

This strategy was applied  other weather variables, such as solar radiation, humidity, etc. 
For the daily rain, we used before to that a uniform
probability to decide whether it had rained or not, and then proceeded to
calculate the mm of rain according to a random sample from the normal
distribution.

\subsection{Neural-network surrogate model}
\label{surrogateNN}

Applying the sampling method described above, we executed DSSAT several times
to create the training set for the neural network, to be used as
a surrogate.

 The neural network presented a simple architecture with one hidden
layer with 1000 neurons and ReLU as the activation function. All the input parameters were then
normalized between the interval $[0, 1]$, as well as the output yield values
calculated for these parameters by DSSAT.

\subsection{Results}

A surrogate model can be considered a good one if it can replicate similar
results of the target model, with reduced computational time. In total, we have
generated 700 different combinations of parameters with the quasi-random
sampling strategy, and the associated yield calculated by DSSAT. We have user
75\% of this dataset as the training set, using the remaining 25\% as the test
set for validation purposes. Our approach with NN achieved satisfactory results,
presenting an average $R^2$ correlation equal to $0.96$ (obtained from 5 rounds
of execution), and an evident computational time reduction. To execute DSSAT for
the training and test datasets ($700$ entries in total), it was required 450
seconds on average to complete the task on a single core machine with $16$GB
RAM. Test dataset evaluation ($175$ entries) took $96$ seconds. On the other
hand, the surrogate model running on the same machine required less than a
second to perform the forecast for the whole test dataset ($175$ entries).
It is clear that regarding time and computational consumption there was a clear
improvement with the usage of the surrogate model.

\begin{figure}[!h]
  \centering
  \includegraphics[width=.98\linewidth]{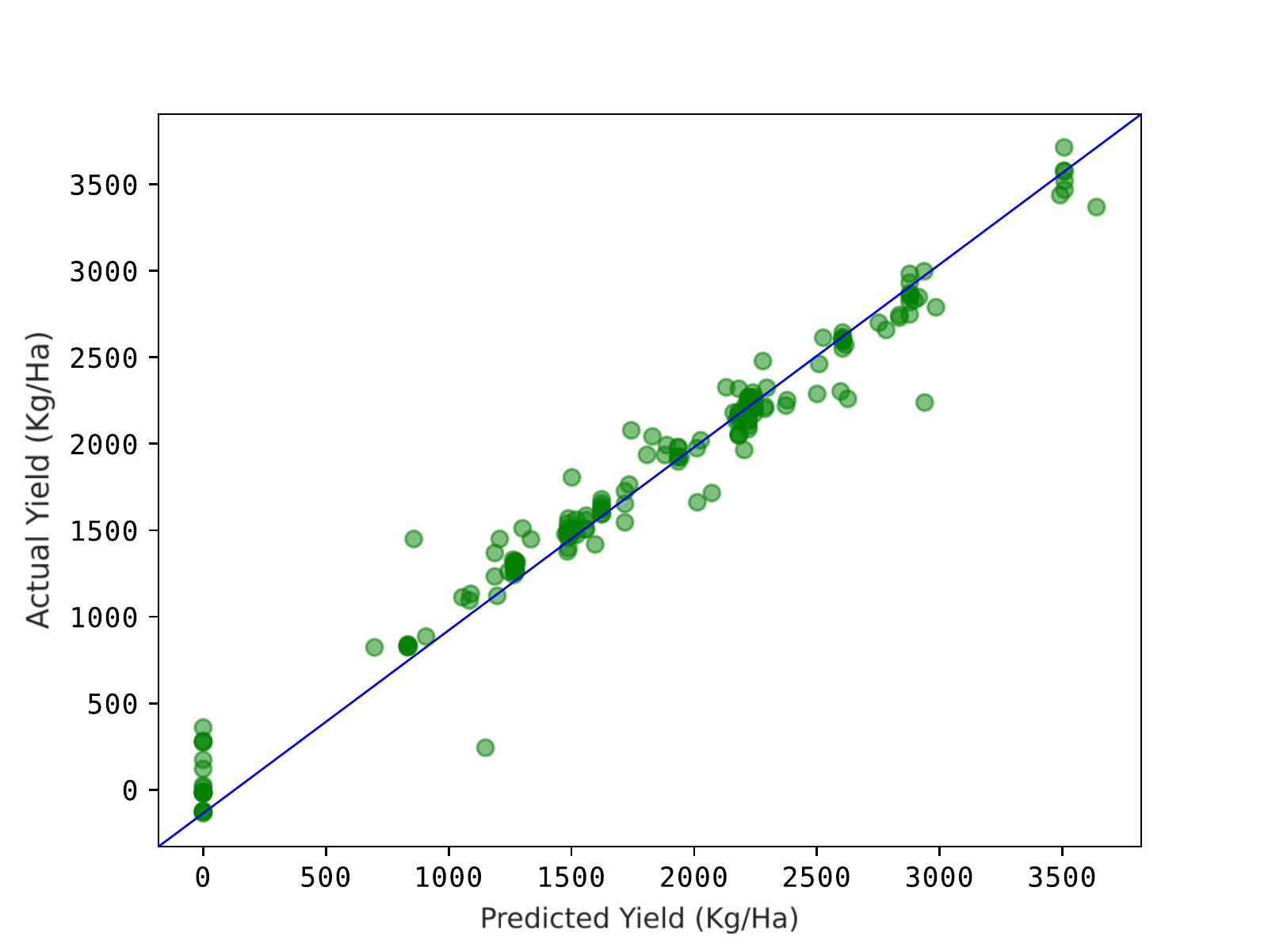}
  \caption{%
    Graphical comparison between data simulated on DSSAT(x axis) and generated by our surrogate model
(y axis). We can se that the results obtained by both models. Minor discrepancies, specially 
in the extremes can be disregarded by defining a small confidence interval where the surrogate
model can replace the crop simulation model with adequate accuracy.
    }\label{fig:nn-dssat}
\end{figure}

The high accuracy in the prediction provided by the NN can indicate possible
overfitting of data, which would demand validation with a totally different
test set. The test set was composed of random entries generated by the
quasi-random method over a specified range of each parameter. So, it is very likely
that each sample from the test set presented a very close entry used in the
training phase. It indicates, however, that performing an extensive and
efficient sampling of the parameters space to generate artificial data for the
construction of an NN-based surrogate is promising. If we can cover reasonably well the whole
parameter space for the construction of the surrogate model, chances are high
that a new entrance is somehow close to one instance in the training set. In
this sense, the use of the quasi-random method for sampling seems to be the best
option, as it mixes random samples and guided samples to unexplored areas of the
search space.

In the future, this work can be extended in a way to better estimate the optimal
range of variation of the input parameters and the minimum number of samples 
to construct a surrogate with a target accuracy.

\section{Conclusion}

This work presented a comprehensive method to predict crop yield using different
approaches that leverage both crop simulation and data-driven models attending
to multiple objectives and user needs. We detail a data-driven crop yield model
capable of predicting yield and providing related probability distribution
functions to risk assessment tasks.  Our data-driven results outperformed the
previous work, with a correlation of 91\% for corn and 45\% for soybean,
 while also providing uncertain information. Different from
previous work, we presented a machine learning-based model to estimate yield for
a given region and provide the underlying probability distribution function for
risk assessment activities.  Using these probability distribution functions
farmers and food security practitioners can estimate risk metrics like the
probability of worst-case scenarios under certain weather conditions instead of
single-point estimates like most of the current works do.

Additionally, we described a crop simulation model calibration and evaluation
system that leverages high-performance computing to predict crop yield for large
areas and support food security decision-making.  The presented crop simulation
model infrastructure enables the calibration and simulation of crop growths for
large regions using fine-grained basic simulators.  The crop simulation model
calibration and evaluation infrastructure enable users to create ensembles of
previously calibrated models to enhance predictions giving more weight to
calibrated models that performed better in the past.
Our results showed a year-based crop simulation ensemble model with a 91\%
correlation between mean predicted and measured values, and a 6\% mean absolute
percentage error for soybean in Illinois (USA).

 The presented surrogate crop simulation model generates a 96\% correlation
 between the actual model execution and their predictions while reducing sharply
 the required  computational resources. Our approach enables a large range of
 decision-making levels including single farm management insights to
 production-level government policies. For users with low computational
 capabilities, we present a lightweight surrogate model to speed up the
 evaluation process at low model performance expenses.  To the best of our
 knowledge, this is the first work that combines data-driven models and crop
 simulation models to enable different users to accomplish their tasks on an
 integrated platform.



\bibliographystyle{apa}
\bibliography{references}

\begin{thebibliography}{}

\bibitem[\protect\astroncite{Akiba et~al.}{2019}]{optuna_2019}
Akiba, T., Sano, S., Yanase, T., Ohta, T., and Koyama, M. (2019).
\newblock Optuna: A next-generation hyperparameter optimization framework.
\newblock In {\em Proceedings of the 25rd {ACM} {SIGKDD} International
  Conference on Knowledge Discovery and Data Mining}.

\bibitem[\protect\astroncite{Akinseye et~al.}{2017}]{akinseye2017assessing}
Akinseye, F.~M., Adam, M., Agele, S.~O., Hoffmann, M.~P., Traore, P., and
  Whitbread, A.~M. (2017).
\newblock Assessing crop model improvements through comparison of sorghum
  (sorghum bicolor l. moench) simulation models: a case study of west african
  varieties.
\newblock {\em Field Crops Research}, 201:19--31.

\bibitem[\protect\astroncite{Audet and Hare}{2017}]{audet2017derivative}
Audet, C. and Hare, W. (2017).
\newblock {\em Derivative-Free and Blackbox Optimization}.
\newblock Springer Series in Operations Research and Financial Engineering.
  Springer International Publishing.

\bibitem[\protect\astroncite{Bailey et~al.}{2011}]{bailey2011growing}
Bailey, R. et~al. (2011).
\newblock Growing a better future: food justice in a resource-constrained
  world.
\newblock {\em Growing a better future: food justice in a resource-constrained
  world.}

\bibitem[\protect\astroncite{Battisti et~al.}{2017}]{battisti2017inter}
Battisti, R., Sentelhas, P.~C., and Boote, K.~J. (2017).
\newblock Inter-comparison of performance of soybean crop simulation models and
  their ensemble in southern brazil.
\newblock {\em Field crops research}, 200:28--37.

\bibitem[\protect\astroncite{de~Freitas~Cunha and
  Silva}{2020}]{cunha2020estimating}
de~Freitas~Cunha, R.~L. and Silva, B. (2020).
\newblock Estimating crop yields with remote sensing and deep learning.
\newblock In {\em 2020 IEEE Latin American GRSS ISPRS Remote Sensing Conference
  (LAGIRS)}, pages 273--278.

\bibitem[\protect\astroncite{de~Geografia~e
  Estatística~(IBGE)}{2016}]{ibgepma}
de~Geografia~e Estatística~(IBGE), I.~B. (2016).
\newblock Produ\c{c}\~{a}o agr\'{i}cola municipal.
\newblock http://www.ibge.gov.br.

\bibitem[\protect\astroncite{Dee et~al.}{2011}]{dee2011era}
Dee, D.~P., Uppala, S., Simmons, A., Berrisford, P., Poli, P., Kobayashi, S.,
  Andrae, U., Balmaseda, M., Balsamo, G., Bauer, d.~P., et~al. (2011).
\newblock {The ERA-Interim reanalysis: Configuration and performance of the
  data assimilation system}.
\newblock {\em Quarterly Journal of the royal meteorological society},
  137(656):553--597.

\bibitem[\protect\astroncite{Frangos et~al.}{2010}]{frangos2010surrogate}
Frangos, M., Marzouk, Y., Willcox, K., and van Bloemen~Waanders, B. (2010).
\newblock Surrogate and reduced-order modeling: a comparison of approaches for
  large-scale statistical inverse problems [chapter 7].

\bibitem[\protect\astroncite{Funk et~al.}{2015}]{funk2015climate}
Funk, C., Peterson, P., Landsfeld, M., Pedreros, D., Verdin, J., Shukla, S.,
  Husak, G., Rowland, J., Harrison, L., Hoell, A., et~al. (2015).
\newblock The climate hazards infrared precipitation with stations—a new
  environmental record for monitoring extremes.
\newblock {\em Scientific data}, 2:150066.

\bibitem[\protect\astroncite{Gal and Ghahramani}{2016}]{gal2016dropout}
Gal, Y. and Ghahramani, Z. (2016).
\newblock Dropout as a bayesian approximation: Representing model uncertainty
  in deep learning.
\newblock In {\em international conference on machine learning}, pages
  1050--1059.

\bibitem[\protect\astroncite{George et~al.}{1990}]{george1990effect}
George, T., Bartholomew, D.~P., Singleton, P.~W., et~al. (1990).
\newblock Effect of temperature and maturity group on phenology of field grown
  nodulating and nonnodulating soybean isolines.
\newblock {\em Biotronics}, 19:49--59.

\bibitem[\protect\astroncite{Godfray and Garnett}{2014}]{godfray2014food}
Godfray, H. C.~J. and Garnett, T. (2014).
\newblock Food security and sustainable intensification.
\newblock {\em Philosophical transactions of the Royal Society B: biological
  sciences}, 369(1639):20120273.

\bibitem[\protect\astroncite{Han et~al.}{2012}]{han2012cropscape}
Han, W., Yang, Z., Di, L., and Mueller, R. (2012).
\newblock Cropscape: A web service based application for exploring and
  disseminating us conterminous geospatial cropland data products for decision
  support.
\newblock {\em Computers and Electronics in Agriculture}, 84:111--123.

\bibitem[\protect\astroncite{Hengl et~al.}{2017}]{hengl2017soilgrids250m}
Hengl, T., de~Jesus, J.~M., Heuvelink, G.~B., Gonzalez, M.~R., Kilibarda, M.,
  Blagoti{\'c}, A., Shangguan, W., Wright, M.~N., Geng, X.,
  Bauer-Marschallinger, B., et~al. (2017).
\newblock Soilgrids250m: Global gridded soil information based on machine
  learning.
\newblock {\em PLoS one}, 12(2):e0169748.

\bibitem[\protect\astroncite{Hochreiter and
  Schmidhuber}{1997}]{hochreiter1997long}
Hochreiter, S. and Schmidhuber, J. (1997).
\newblock Long short-term memory.
\newblock {\em Neural computation}, 9(8):1735--1780.

\bibitem[\protect\astroncite{Hoogenboom et~al.}{2019}]{hoogenboom2019dssat}
Hoogenboom, G., Porter, C., Boote, K., Shelia, V., Wilkens, P., Singh, U.,
  White, J., Asseng, S., Lizaso, J., Moreno, L., et~al. (2019).
\newblock The dssat crop modeling ecosystem.
\newblock {\em Advances in crop modelling for a sustainable agriculture}, pages
  173--216.

\bibitem[\protect\astroncite{Jones and Johnson}{2009}]{jones2009design}
Jones, B. and Johnson, R.~T. (2009).
\newblock Design and analysis for the gaussian process model.
\newblock {\em Quality and Reliability Engineering International},
  25(5):515--524.

\bibitem[\protect\astroncite{Keating et~al.}{2014}]{keating2014food}
Keating, B.~A., Herrero, M., Carberry, P.~S., Gardner, J., and Cole, M.~B.
  (2014).
\newblock Food wedges: framing the global food demand and supply challenge
  towards 2050.
\newblock {\em Global Food Security}, 3(3-4):125--132.

\bibitem[\protect\astroncite{Kim and Lee}{2016}]{kim2016machine}
Kim, N. and Lee, Y.-W. (2016).
\newblock Machine learning approaches to corn yield estimation using satellite
  images and climate data: a case of iowa state.
\newblock {\em Journal of the Korean Society of Surveying, Geodesy,
  Photogrammetry and Cartography}, 34(4):383--390.

\bibitem[\protect\astroncite{Kucherenko et~al.}{2015}]{kucherenko2015exploring}
Kucherenko, S., Albrecht, D., and Saltelli, A. (2015).
\newblock Exploring multi-dimensional spaces: A comparison of latin hypercube
  and quasi monte carlo sampling techniques.
\newblock {\em arXiv preprint arXiv:1505.02350}.

\bibitem[\protect\astroncite{Kuleshov et~al.}{2018}]{kuleshov2018accurate}
Kuleshov, V., Fenner, N., and Ermon, S. (2018).
\newblock Accurate uncertainties for deep learning using calibrated regression.
\newblock In {\em International Conference on Machine Learning}, pages
  2796--2804. PMLR.

\bibitem[\protect\astroncite{Kuwata and Shibasaki}{2015}]{kuwata2015estimating}
Kuwata, K. and Shibasaki, R. (2015).
\newblock Estimating crop yields with deep learning and remotely sensed data.
\newblock In {\em 2015 IEEE International Geoscience and Remote Sensing
  Symposium (IGARSS)}, pages 858--861. IEEE.

\bibitem[\protect\astroncite{Lobell et~al.}{2009}]{lobell2009crop}
Lobell, D.~B., Cassman, K.~G., and Field, C.~B. (2009).
\newblock Crop yield gaps: their importance, magnitudes, and causes.
\newblock {\em Annual review of environment and resources}, 34.

\bibitem[\protect\astroncite{Mendes et~al.}{2019}]{mendes2019calendario}
Mendes, A.~D., Sousa, T. D. G.~F., and Mendes, N.~P. (2019).
\newblock {Calendario de Plantio e Colheita de grãos no Brasil}.
\newblock Technical report, Companhia Nacional de Abastecimento (Conab).

\bibitem[\protect\astroncite{NASS}{2011}]{nass2011quick}
NASS, U. (2011).
\newblock Quick stats 2.0.
\newblock {\em United States Department of Agriculture National Agricultural
  Statistics Service}.

\bibitem[\protect\astroncite{NASS}{2018}]{nass2018united}
NASS, U. (2018).
\newblock United states department of agriculture-national agricultural
  statistics service.
\newblock {\em Cropland Data Layer}.

\bibitem[\protect\astroncite{Nearing et~al.}{2012}]{nearing2012}
Nearing, G., Crow, W., Thorp, K., Moran, M., Reichle, R., and Gupta, H. (2012).
\newblock Assimilating remote sensing observations of leaf area index and soil
  moisture for wheat yield estimates: An observing system simulation
  experiment.
\newblock {\em Water Resources Research}, 48.

\bibitem[\protect\astroncite{Neild and Newman}{1990}]{neild1987nch}
Neild, R.~E. and Newman, J.~E. (1990).
\newblock Nch-40 growing season characteristics and requirements in the corn
  belt.

\bibitem[\protect\astroncite{Nguyen et~al.}{2019}]{nguyen2019spatial}
Nguyen, L.~H., Zhu, J., Lin, Z., Du, H., Yang, Z., Guo, W., and Jin, F. (2019).
\newblock Spatial-temporal multi-task learning for within-field cotton yield
  prediction.
\newblock In Yang, Q., Zhou, Z.-H., Gong, Z., Zhang, M.-L., and Huang, S.-J.,
  editors, {\em Advances in Knowledge Discovery and Data Mining}, pages
  343--354, Cham. Springer International Publishing.

\bibitem[\protect\astroncite{Oliveira et~al.}{2018}]{oliveira2018scalable}
Oliveira, I., Cunha, R. L.~F., Silva, B., and Netto, M. A.~S. (2018).
\newblock A scalable machine learning system for pre-season agriculture yield
  forecast.
\newblock In {\em 2018 IEEE 14th International Conference on e-Science
  (e-Science)}, pages 423--430.

\bibitem[\protect\astroncite{Rodriguez et~al.}{2018}]{rodriguez2018predicting}
Rodriguez, D., De~Voil, P., Hudson, D., Brown, J., Hayman, P., Marrou, H., and
  Meinke, H. (2018).
\newblock Predicting optimum crop designs using crop models and seasonal
  climate forecasts.
\newblock {\em Scientific reports}, 8(1):1--13.

\bibitem[\protect\astroncite{Saltelli et~al.}{2010}]{saltelli2010variance}
Saltelli, A., Annoni, P., Azzini, I., Campolongo, F., Ratto, M., and Tarantola,
  S. (2010).
\newblock Variance based sensitivity analysis of model output. design and
  estimator for the total sensitivity index.
\newblock {\em Computer physics communications}, 181(2):259--270.

\bibitem[\protect\astroncite{Silva et~al.}{2018}]{silva2018jobpruner}
Silva, B., Netto, M.~A., and Cunha, R.~L. (2018).
\newblock Jobpruner: A machine learning assistant for exploring parameter
  spaces in hpc applications.
\newblock {\em Future Generation Computer Systems}, 83:144--157.

\bibitem[\protect\astroncite{Solomatine et~al.}{2009}]{solomatine2009data}
Solomatine, D., See, L.~M., and Abrahart, R. (2009).
\newblock Data-driven modelling: concepts, approaches and experiences.
\newblock {\em Practical hydroinformatics}, pages 17--30.

\bibitem[\protect\astroncite{Sultan et~al.}{2019}]{sultan2019evidence}
Sultan, B., Defrance, D., and Iizumi, T. (2019).
\newblock Evidence of crop production losses in west africa due to historical
  global warming in two crop models.
\newblock {\em Scientific reports}, 9(1):1--15.

\bibitem[\protect\astroncite{Sun and Sun}{2015}]{sun2015model}
Sun, N.-Z. and Sun, A. (2015).
\newblock {\em Model calibration and parameter estimation: for environmental
  and water resource systems}.
\newblock Springer.

\bibitem[\protect\astroncite{Tripathy and Bilionis}{2018}]{tripathy2018deep}
Tripathy, R.~K. and Bilionis, I. (2018).
\newblock Deep uq: Learning deep neural network surrogate models for high
  dimensional uncertainty quantification.
\newblock {\em Journal of computational physics}, 375:565--588.

\bibitem[\protect\astroncite{{van Klompenburg}
  et~al.}{2020}]{vanklompenburg2020105709}
{van Klompenburg}, T., Kassahun, A., and Catal, C. (2020).
\newblock Crop yield prediction using machine learning: A systematic literature
  review.
\newblock {\em Computers and Electronics in Agriculture}, 177:105709.

\bibitem[\protect\astroncite{Vladimirova
  et~al.}{2019}]{vladimirova2019understanding}
Vladimirova, M., Verbeek, J., Mesejo, P., and Arbel, J. (2019).
\newblock Understanding priors in bayesian neural networks at the unit level.
\newblock In {\em International Conference on Machine Learning}, pages
  6458--6467.

\bibitem[\protect\astroncite{Zhang et~al.}{2021}]{zhang2021408}
Zhang, S., Bai, Y., hua Zhang, J., and Ali, S. (2021).
\newblock Developing a process-based and remote sensing driven crop yield model
  for maize (prym–maize) and its validation over the northeast china plain.
\newblock {\em Journal of Integrative Agriculture}, 20(2):408--423.

\end{thebibliography}

\end{document}